\documentclass[a4paper,fleqn]{cas-dc}

\usepackage[authoryear]{natbib}

\begin{document}
\let\WriteBookmarks\relax
\def\floatpagepagefraction{1}
\def\textpagefraction{.001}

\title[mode = title]{T-UNet: Triplet UNet for Change Detection in High-Resolution Remote Sensing Images}


\author[1]{Huan Zhong}
\ead{zhonghuan@whu.edu.cn}
\address[1]{State Key Laboratory of Information Engineering in Surveying, Mapping and Remote Sensing, Wuhan University, Wuhan 430079, China}

\author[1]{Chen Wu}
\ead{chen.wu@whu.edu.cn}
\cormark[1] 
\cortext[1]{Corresponding author}

\begin{abstract}[S U M M A R Y]
Remote sensing image change detection aims to identify the differences between images acquired at different times in the same area. It is widely used in land management, environmental monitoring, disaster assessment and other fields. Currently, most change detection methods are based on Siamese network structure or early fusion structure. Siamese structure focuses on extracting object features at different times but lacks attention to change information, which leads to false alarms and missed detections. Early fusion (EF) structure focuses on extracting features after the fusion of images of different phases but ignores the significance of object features at different times for detecting change details, making it difficult to accurately discern the edges of changed objects. To address these issues and obtain more accurate results, we propose a novel network, Triplet UNet(T-UNet), based on a three-branch encoder, which is capable to simultaneously extract the object features and the change features between the pre- and post-time-phase images through triplet encoder. To effectively interact and fuse the features extracted from the three branches of triplet encoder, we propose a multi-branch spatial-spectral cross-attention module (MBSSCA). In the decoder stage, we introduce the channel attention mechanism (CAM) and spatial attention mechanism (SAM) to fully mine and integrate detailed textures information at the shallow layer and semantic localization information at the deep layer. The proposed network T-UNet is compared with seven other state-of-the-art change detection methods on three publicly available datasets. Experimental results show that the proposed T-UNet achieves the best results in terms of comprehensive evaluation metrics on all three datasets, significantly outperforming other comparative methods. Extensive experiments verify the effectiveness of the proposed structure and modules as well as the superiority of the proposed T-UNet. The source code of the proposed T-UNet is available at \href{https://github.com/Pl-2000/T-UNet}{https://github.com/Pl-2000/T-UNet}.
\end{abstract}


\begin{highlights}
	\item highlights1
This paper proposes a novel change detection network structure, named Triplet UNet (T-UNet). It addresses the drawbacks of the EF and Siamese structures, and represents a new attempt in the field of CD.
	\item highlights2
Multi-Branch Spatial-Spectral Cross Attention (MBSSCA) module is designed to interact and fuse the features extracted from the triplet encoder with three branches.
	\item highlights3
Extensive experiments on three publicly available change detection datasets have confirmed the superior performance of the proposed network T-UNet and validated the necessity and effectiveness of the proposed structures and modules.
\end{highlights}

\maketitle

\section{Introduction}
Remote sensing (RS) image change detection (CD) refers to the technique of extracting information about objects from two or more images acquired in the same geographical area but at different times, and then detecting whether or not and what kind of changes have occurred in the geometry or spectral properties of the target objects\citep{ref1,ref2}. Change detection of bi-temporal remote sensing images has been a hot research topic in the RS community, and it has been widely applied in various scenarios, such as land management\citep{ref3,ref4}, environmental monitoring\citep{ref5}, disaster assessment\citep{ref6,ref7,ref8}, and urban development analysis\citep{ref9}. With the development of remote sensing technology, the sources of remote sensing data have become increasingly diverse, the image resolution has become higher, and the image details have become richer\citep{ref10,ref11}. Recently, high-resolution remote sensing images with rich spatial details have become the main data source for change detection, especially suitable for fine-grained scenarios such as urbanization development change detection.

\par In the past few decades, research on change detection algorithms has been changing rapidly. This paper categorizes these methods into two major groups: traditional change detection methods and deep learning-based methods. Traditional change detection methods mainly analyze the spectral information of the images and select an appropriate threshold to partition the change regions, including image algebra\citep{ref12,ref13}, change vector analysis (CVA)\citep{ref14,ref15}, multivariate alteration detection (MAD)\citep{ref16}, principal component analysis (PCA)\citep{ref17}, and slow feature analysis (SFA)\citep{ref18,ref19}. Among them, the most commonly used threshold selection methods are Otsu thresholding \citep{ref20}and Kullback-Leibler divergence method\citep{ref21}. However, these methods are susceptible to the data distribution due to their simplicity and straightforwardness. As machine learning (ML) develops, a large number of related strategies have been proposed for feature analysis and change detection, including support vector machine (SVM)\citep{ref22,ref23}, decision tree\citep{ref24}, and random forest\citep{ref25}. However, these ML-based methods require manual feature design and extraction, making it difficult to filter and extract the critical features from rich and complex information in high-resolution remote sensing images. Moreover, manual feature extraction heavily relies on expertise, greatly affecting the automation capability of CD technology. Overall, the limitations of traditional change detection methods are becoming increasingly apparent.

\par In recent years, deep learning (DL) algorithms have received widespread attention for their powerful representation learning capabilities. Deep learning has achieved success in various fields such as computer vision\citep{ref26,ref27}, natural language processing\citep{ref28}, information retrieval\citep{ref29}, and recommendation systems\citep{ref30}. Compared with traditional methods, DL-based methods can learn and reproduce high-dimensional features using deep structures without the need for manual feature design or rule setting, significantly reducing the demand for expertise. Researchers are increasingly turning their attention to deep learning techniques, including convolutional neural network (CNN)\citep{ref31,ref32,ref33}, recurrent neural network (RNN)\citep{ref34,ref35}, visual transformer (ViT)\citep{ref36,ref37}, and graph neural network (GNN)\citep{ref38,ref39}, and have developed a series of DL-based change detection networks to improve accuracy and its automation capability.

\par The DL-based CD methods can be divided into two categories: patch-based methods and image-based methods. The patch-based method considers the neighborhood of the pixel to be detected as a whole patch for feature extraction and analysis. This method requires separate analysis and prediction for the neighborhood of each pixel, which is extremely inefficient. Currently, the mainstream DL-based methods are image-based, as they can achieve end-to-end dense prediction efficiently, i.e., image-to-image prediction. The most typical image-based framework is UNet, a variant of the fully convolutional neural network (FCN) framework\citep{ref40}. FCN uses the fused features of the bi-temporal images through concatenation or difference as input and outputs the change map. Compared with FCN, UNet uses skip-connection structure to bridge the information at different levels between encoder and decoder to enhance feature representation. Later, residual structures are introduced to further deepen the network and improve the feature extraction capability of the network\citep{ref41,ref42}. The most straightforward approach of UNet change detection methods processing multi-temporal images is early fusion (EF) structure, whose encoder only contains one branch and the multi-temporal images are concatenated as the input. Premature fusion of the bi-temporal images by this structure ignores the original object features and thus loses the crucial semantic information of the pre- and post-time-phase images, bringing in noise and generating pseudo changes.

\par To this end, Daudt et al. proposed a Siamese network with a dual-branch structure to input bi-temporal images for change detection by comparing different hierarchical features of the original images\citep{ref43}. To maintain the features extracted from the two branches in the same feature space, the two branches often use sub-networks with the identical structure and weight sharing. The Siamese structure effectively improves the performance of the CD models. Since then, this idea has been widely adopted, and many variants based on the Siamese UNet have sprung up\citep{ref44,ref45,ref46,ref47,ref48,ref49,ref50}. These variants use the Siamese network structure as the encoder for feature extraction, and the decoder for feature analysis and change detection. \citet{ref45} used the Siamese network structure in the encoder phase to extract features of the bi-temporal images, and a deep supervision module in the decoder phase to enhance the representation and recognition capability. Chen et al. \citep{ref51}embedded a dual-attention module in the Siamese network to extract high-dimensional features of the bi-temporal images for similarity calculation. \citet{ref50} proposed a difference enhancement module and embedded it in the Siamese encoder to enhance the focus on the changed areas. \citet{ref49} used Siamese sub-networks as encoders to extract deep features of the original bi-temporal images and used an ensemble channel attention module to aggregate and refine semantic features at different levels. The methods mentioned above can be classified as Siamese-based methods, whose encoder contains two branches. These Siamese UNet-based methods improve the accuracy of CD effectively, but they still have some noticeable drawbacks. The Siamese structure greatly preserves the object features of the pre- and post-time-phase images, but it cannot fully utilize the change information, resulting in inaccurate edge detection of change regions, and producing more false detections and misses.

\par To address the aforementioned issues, a novel change detection network, Triplet UNet (T-UNet) of encoder with three branches, is proposed in this paper. To the best of my knowledge, this paper is the first to propose a three-branch network framework for CD tasks. The proposed T-UNet consists of two stages, the encoder and decoder stages. Unlike EF and Siamese network structures, the encoder of the proposed network contains three branches. One is the main branch of the differential image data stream, which is used to input the differential image of the bi-temporal images to be detected. The other two branches are the auxiliary branches of the original image data stream, which are used to input each of the bi-temporal images. In this way, the proposed network can simultaneously extract and synthesize the object features of the pre- and post-time-phase images and change features between them. This can concurrently solve the shortcomings of the EF and Siamese structures. To effectively integrate the features extracted from the three branches, we propose a Multi-Branch Spatial-Spectral Cross Attention (MBSSCA) module. The MBSSCA module identifies and enhances the change features in different branches from the spectral and spatial dimensions respectively, and then uses the attention-enhanced change intensity maps to filter and transform the key information, thus effectively integrating the critical features of the three branches. Additionally, the change feature maps obtained from MBSSCA in the main branch encoder are passed to the corresponding levels of the decoder through skip connection structures, and thus used for complementary fusion of the change features of various coarse and fine grain. In the decoder stage, the channel attention mechanism (CAM) and spatial attention mechanism (SAM) are introduced to fully exploit and integrate detail textures at shallow layers and semantic localization information at deep layers, providing accurate and robust support for detecting changes. In the end, the sigmoid function is utilized to process the final feature map to obtain the change probability map, and the change detection result is obtained by thresholding.

\par In summary, the main contributions of this paper are summarized as follows.

\begin{enumerate}[1)]
	
	\item{This paper proposes a novel change detection network structure, named Triplet UNet (T-UNet). With its encoder containing three branches, it can simultaneously extract and utilize the object features of the pre- and post-time-phase images and their corresponding change features, thus reducing the loss of critical information during feature extraction. This addresses the drawbacks of the EF and Siamese structures, and represents a new attempt in the field of CD.}
	
	\item{Multi-Branch Spatial-Spectral Cross Attention (MBSSCA) module is designed to interact and fuse the features extracted from the triplet encoder with three branches. It leverages the detail features of the bi-temporal images to compare and correct the change features in the branch of the differential image data stream. The correction is twofold: firstly, identifying and suppressing pseudo changes caused by direct differential calculation of the bi-temporal images or noise; secondly, checking and supplementing missing real change features. Additionally, the MBSSCA module is capable of recognizing and enhancing change features in both spectral and spatial dimensions, as well as filtering and transforming key information from the three branches.}
	
	\item{Extensive experiments on three publicly available change detection datasets have confirmed the superior performance of the proposed network T-UNet and validated the necessity and effectiveness of the proposed structures and modules. Experimental results demonstrate that the proposed T-UNet achieves the best F1 and Overall Accuracy (OA) on the LEVIR-CD, WHU-CD, and DSIFN-CD datasets, significantly outperforming the other seven comparative methods.}
	
\end{enumerate}

\par The remainder of this paper is organized as follows. Section \ref{Methodology} provides a detailed introduction of our proposed T-UNet network and its respective modules. Section \ref{Experimental Setting} describes the datasets, compared methods, and evaluation metrics involved in the experiments. Section \ref{Experimental Results and Discussion} then delivers a detailed presentation and analysis of the experimental results. Finally, we discuss and summarize this paper in Section \ref{Conclusion}.

\section{Methodology}
\label{Methodology}

\par In this section, we first present the overall framework of the proposed network T-UNet. The key structures and modules are then described in detail, including the triplet encoder, the multi-branch spatial-spectral cross-attention module and the decoder. Finally, we introduce the loss function used during training stage.

\subsection{Framework Overview}

The existing UNet-based CD networks can be divided into two main types: Early Fusion (EF) and Siamese (Siam) structures. The EF structure contains only one encoder, and the bi-temporal images to be detected are first fused through operations such as concatenation or difference, and then used as input to the network for feature extraction and change detection. This structure can retain rich change features, but the early fusion of images may ignore the object features with key semantic information of the pre- and post-time-phase images, resulting in noise and pseudo-change. The Siamese structure usually consists of two encoders with the identical structure and shared weights, and the input is the bi-temporal images at different moments to be detected. This structure preserves the object features of the pre- and post-time-phase images to a great extent, but cannot fully utilize the change information, thus making the edge detection of changed regions out of place and resulting in more false alarms and missed detections. Therefore, we propose a CD network framework based on a three-branch UNet to simultaneously extract and comprehensively utilize both the object features of the pre- and post-time-phase images and the change features, thereby mining more discriminative features for change detection.

\begin{figure*}[!t]
	\centering
	\includegraphics[width=6.6in]{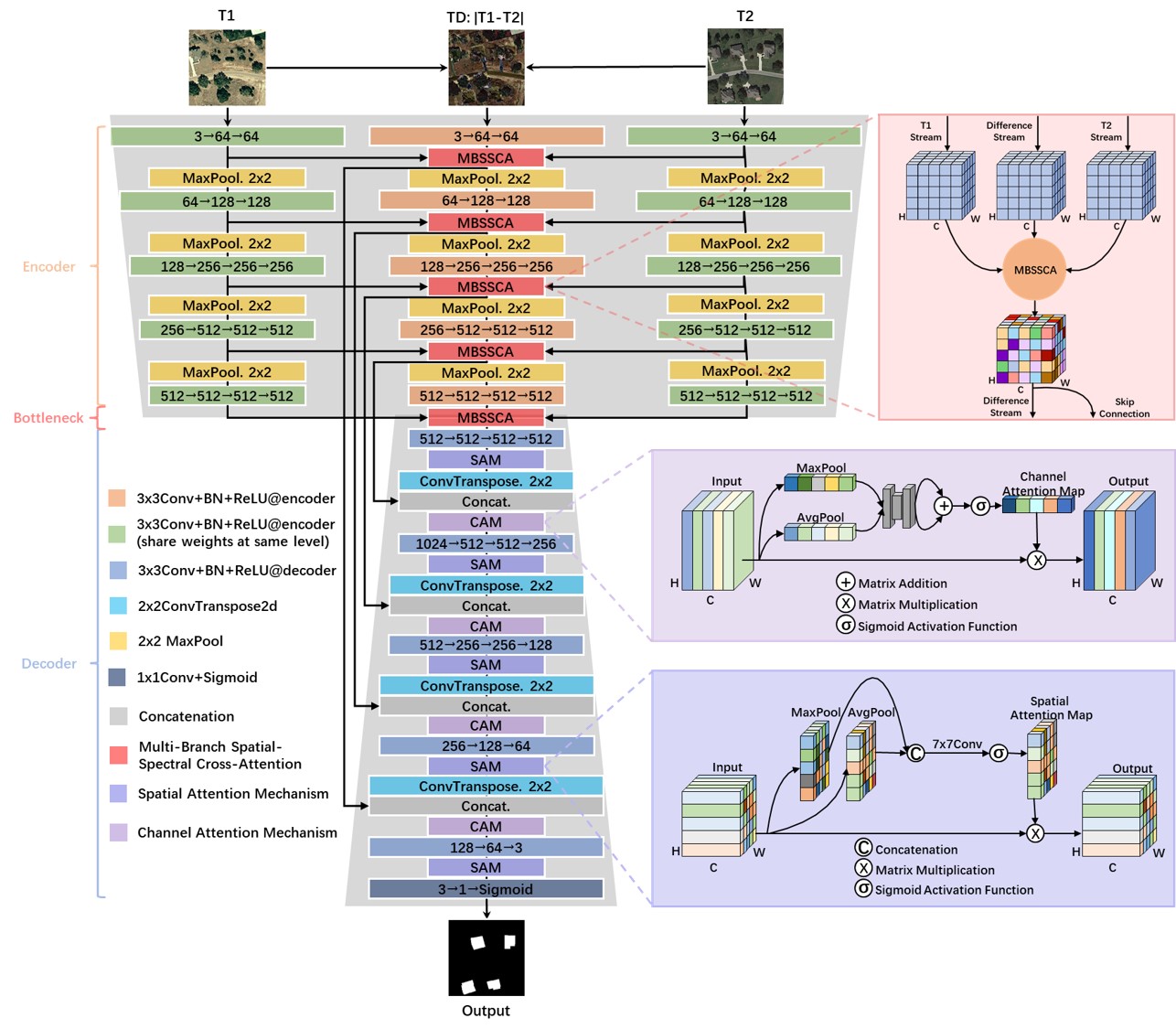}
	\caption{The network structure of the proposed T-UNet.}
	\label{fig1}
\end{figure*}
In this section, we first introduce a novel CD network framework proposed in this paper, T-UNet, and then describe the proposed modules in detail in the following subsections. The proposed network is shown in Fig. \ref{fig1}, which consists of two stages, encoder and decoder. Unlike the EF and Siamese network structures, the encoder of the proposed network contains three branches: a main branch TD for the differential image data stream, which is used to input the differential image of the bi-temporal images to be detected; and two auxiliary branches T1 and T2 for the original image data stream, which are used to input the pre- and post-time-phase images, respectively. The TD branch is used for change information extraction and detection, and the T1 and T2 branches are used for object feature extraction from the pre- and post-time-phase images. To effectively integrate the feature information of the three branches, including the original object features and differential change features, we propose a multi-branch spatial-spectral cross attention (MBSSCA) module. MBSSCA identifies and enhances change features in different branches from both the spectral and spatial dimensions, and then effectively selects and transforms the key features of the three branches using the change intensity map enhanced by cross-attention. In addition, the change feature maps obtained by MBSSCA at different levels of the main branch are transmitted to the corresponding levels of the decoder stage through skip-connection for complementary fusion of different coarse- and fine-grained change features. In the decoder stage, we introduce channel attention mechanism (CAM) and spatial attention mechanism (SAM) to fully exploit and integrate shallow detailed texture information and deep semantic localization information, providing comprehensive and accurate change features for change detection. Finally, the change probability map is obtained with sigmoid function, and the final change detection result is obtained by thresholding. The preset threshold value in the paper is 0.5, i.e. greater than 0.5 is classified as a change category and vice versa. The key modules of the proposed network T-UNet will be elaborated in detail below.

\subsection{Triplet Encoder}

\begin{figure}[!t]
	\centering
	\includegraphics[width=2.8in]{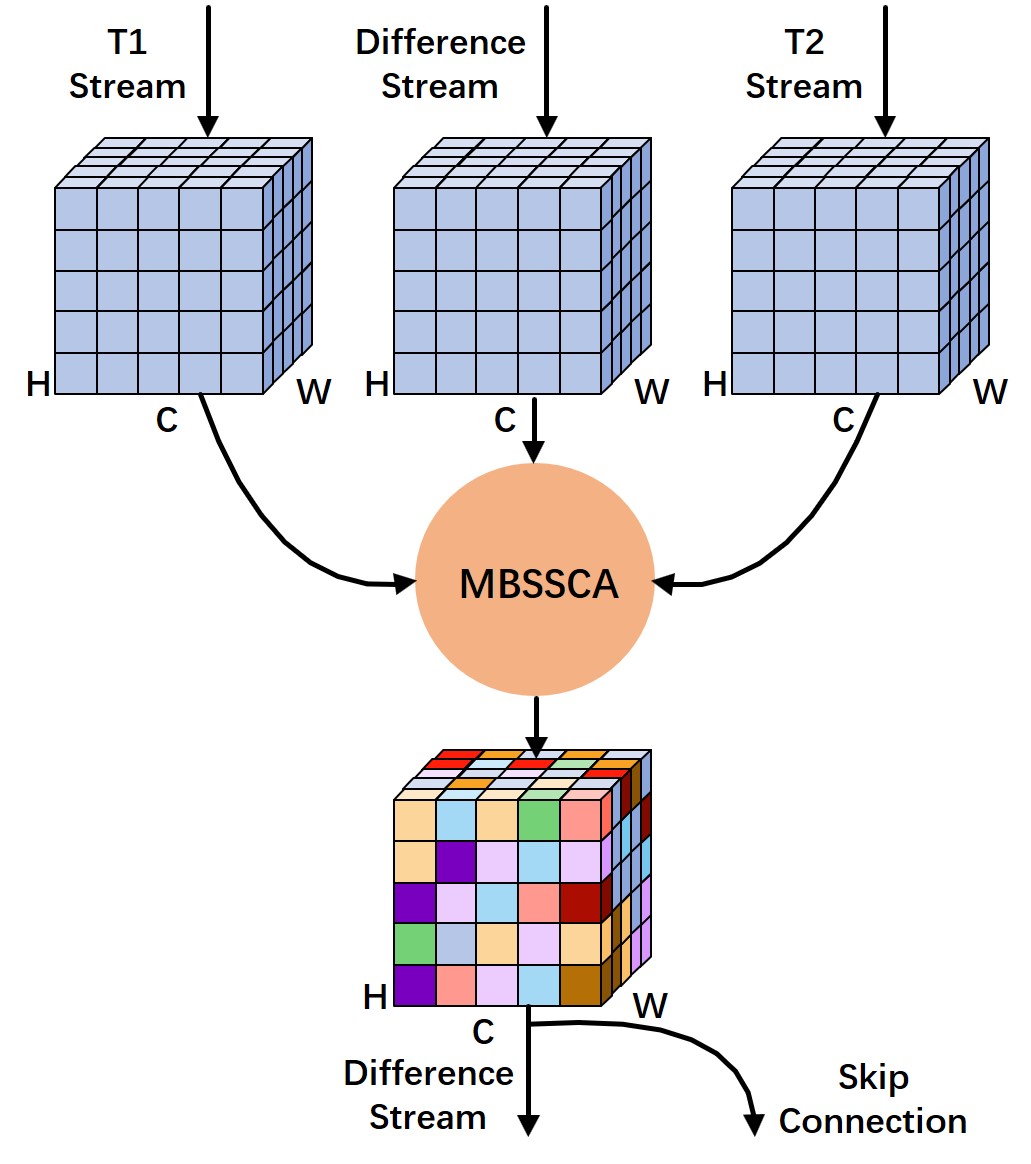}
	\caption{Structure of Triplet Encoder.}
	\label{fig2}
\end{figure}

Fig. \ref{fig2} illustrates the triplet encoder structure of the proposed network, which comprises three branches: the main branch TD for differential image data stream, the T1 branch for the pre-time-phase image data stream, and the T2 branch for the post-time-phase image data stream. These three branches are used to extract the original object features and their differential change features between the pre- and post-time phases. At different levels of the TD branch in the differential image data stream, we propose the MBSSCA module to explore the relationship between the change features and the original object features. By utilizing the detailed texture features of the bi-temporal images to correct the false change information in the change features, the MBSSCA module enhances the representation of the true change features. The features of the three branches are fused through the MBSSCA module to highlight the change features. On the one hand, these features are used as inputs to the TD branch to further explore more representative semantic change features at deeper levels. On the other hand, they are transmitted to the corresponding levels of the decoder stage through skip-connection for complementary fusion of change features at different scales of granularity.

For the construction of the triplet encoder, we selected the modules ahead of the pool5 layer of VGG16 as the backbone of the three-branch sub-networks, as the excellent performance of VGG16 for image feature extraction has been widely proven and recognized. The three sub-networks differ in the specific implementation details. Among them, the sub-networks of T1 and T2 branches adopt the pre-trained VGG16 on ImageNet for extracting features from the original bi-temporal images. Although the ImageNet dataset consists of natural images that are different from RS images, extensive experiments have demonstrated that pre-trained VGG16 on ImageNet can also perform well in feature extraction from RS images. Due to the similarity of imaging principles, data distribution, and feature space between the bi-temporal RS images to be detected, we use two sub-networks with the identical structure and shared weights to process the bi-temporal RS images. On the one hand, the network can better extract and align the features extracted from the pre- and post-time-phase images; on the other hand, the amount of learning parameters is reduced, which is conducive to improving the efficiency of the network. 

As shown in Fig. \ref{fig1}, the sub-network for the T1 and T2 branches consists of five convolutional modules and four max pooling layers, each convolutional module consisting of two or three convolution blocks, and each convolution block consisting of a convolution layer, a batch normalization layer, and a ReLU activation function layer. Let $I_1$ and $I_2$ be the inputs of the T1 and T2 branch, respectively, which are the original bi-temporal images. $l_{p,q}^{I_x}(x=1,2)$ represents the output features of the q-th convolution block in the p-th convolutional module in the data stream corresponding to $I_x$, which is also the input features of the q+1-th convolution block in the p-th convolutional module. Specifically, $l_{1,0}^{I_x}=I_x$. The operation of the convolution block can be expressed as follows.
\begin{equation}
	\label{eq1}
	l_{p,q}^{I_x}=ReLU\left(BN\left(Conv_{p,q}\left(l_{p,q-1}^{I_x}\right)\right)\right)
\end{equation}
where $Conv_{p,q}$(·) represents the convolution operation of the q-th convolution block in the p-th convolutional module of the corresponding branch. $BN$(·) and $ReLU$(·) represent batch normalization and ReLU activation function, respectively. The feature transformation of the pooling layer is illustrated in Eq. \ref{eq2}.
\begin{equation}
	\label{eq2}
	l_{p,0}^{I_x}={MaxPool}_{p-1}\left(l_{p-1,Q}^{I_x}\right)
\end{equation}
where $MaxPool_{p-1}$ represents the max pooling of the (p-1)-th layer, which is used to retain the primary features of the image while reducing the feature dimensionality and preventing overfitting to some extent. $Q$ denotes the total number of convolution blocks in the corresponding convolutional module, and $l_{p-1,Q}^{I_x}$ represents the output feature of the last convolution block in the (p-1)-th convolutional module for the branch corresponding to $I_x$. The features of the bi-temporal images $I_1$ and $I_2$ are extracted by sub-networks of the T1 and T2 branches, obtaining shallow texture details at different levels of granularity at the before and after moments as well as deep semantic features. The final feature maps output by the T1 and T2 branches are $l_{5,3}^{I_1}$ and $l_{5,3}^{I_2}$.

The TD branch processes differential images, which differ significantly from the original bi-temporal images and natural images in terms of data distribution and feature space. Therefore, the TD branch is not suitable for sharing weights with the T1 and T2 branches. In addition, the pre-trained VGG16 on ImageNet is not suitable for change feature extraction in the TD branch. However, we still adopt VGG16 as backbone of the TD branch, but embed the MBSSCA module in it to enable the branch to interact with the semantic features of the pre- and post-time-phase images, extracting more accurate and discriminative change features. Let $I_D$ be the input of the TD branch, i.e., the differential image of the bi-temporal images, as shown in Eq. \ref{eq3}.
\begin{equation}
	\label{eq3}
	I_D=\left|I_1-I_2\right|
\end{equation}

Similar to the representation of the T1 and T2 branches, $l_{p,q}^{I_D}$ denotes the output feature of the q-th convolution block of the p-th convolutional module in the TD branch, and also the input feature of the (q+1)-th convolution block of the p-th convolutional module. Specifically, $l_{1,0}^{I_D}$=$I_D$. The convolution operation and representation of the convolution block in the TD branch are shown in Eq. \ref{eq1}, where $I_x$=$I_D$. After each convolutional module, we utilize the proposed MBSSCA module to interact with the feature information of the three branches. On the one hand, the MBSSCA module compares and corrects the changed features in the TD branch by utilizing the different levels of object feature details of the bi-temporal images; on the other hand, the MBSSCA module identifies and enhances the real changed features in the TD branch. The feature transformation of the MBSSCA module is shown in Eq. \ref{eq4}.
\begin{equation}
	\label{eq4}
	l_{MBSSCA_p}^{I_D}=MBSSCA_p\left(l_{p,Q}^{I_1},l_{p,Q}^{I_D},l_{p,Q}^{I_2}\right)
\end{equation}
where $MBSSCA_p$(·) represents the feature transformation of the p-th MBSSCA module, and $l_{p,Q}^{I_x}(x=1,2,D)$ represents the output feature of the last convolution block in the p-th convolutional module in the branch corresponding to $I_x$. The features of the three branches are interactively fused by the p-th MBSSCA module, and then the feature map $l_{MBSSCA_p}^{I_D}$ is obtained. Subsequently, the significant information in the fused feature map is retained and compressed by pooling operation for further deep feature extraction. In addition, the fused feature map is passed to the decoder stage through skip connection for information complementation and change detection. The feature transformation of the pooling layer in the TD branch is shown in Eq. \ref{eq5}.
\begin{equation}
	\label{eq5}
	l_{p,0}^D=MaxPool_{p-1}\left(l_{{MBSSCA}_{p-1}}^{I_D}\right)
\end{equation}

The features are extracted from the bi-temporal images and their differential image by the three branches of triplet encoder separately, and the features at different levels and granularities are interfused by using the MBSSCA module. Finally, we obtain the output feature $l_{MBSSCA_5}^{I_D}$ of triplet encoder. In the next part, the structure of MBSSCA module is elaborated.

\subsection{Multi-Branch Spatial-Spectral Cross Attention Module}

To effectively integrate the feature information extracted from the three branches of the triplet encoder, including the original object features as well as the changed features, we propose a multi-branch spatial-spectral cross attention (MBSSCA) module. As shown in Fig. \ref{fig1}, each convolutional module is followed by an MBSSCA module that interactively fuses the feature information of the three branches. On the one hand, the MBSSCA module utilizes the detail features extracted from the bi-temporal images to compare and correct the changed features in the TD branch. The correction is twofold: firstly, identifying and suppressing pseudo changes caused by direct differential calculation of the bi-temporal images or noise; secondly, checking and supplementing missing real change features. On the other hand, the MBSSCA module can identify and enhance the changed features from both the spectral and spatial dimensions, and then use the attention-enhanced change intensity map in both the spectral and spatial dimensions to filter and transform the feature representations of the three branches, with the real changed features enhanced and the pseudo-change and unchanged features weakened. The interfused features are used as input to the TD branch for further deep feature extraction and mining. In addition, the fused features are passed to the decoder stage through a skip-connection structure for change information supplementation and detection.

\begin{figure*}[!t]
	\centering
	\includegraphics[width=6.6in]{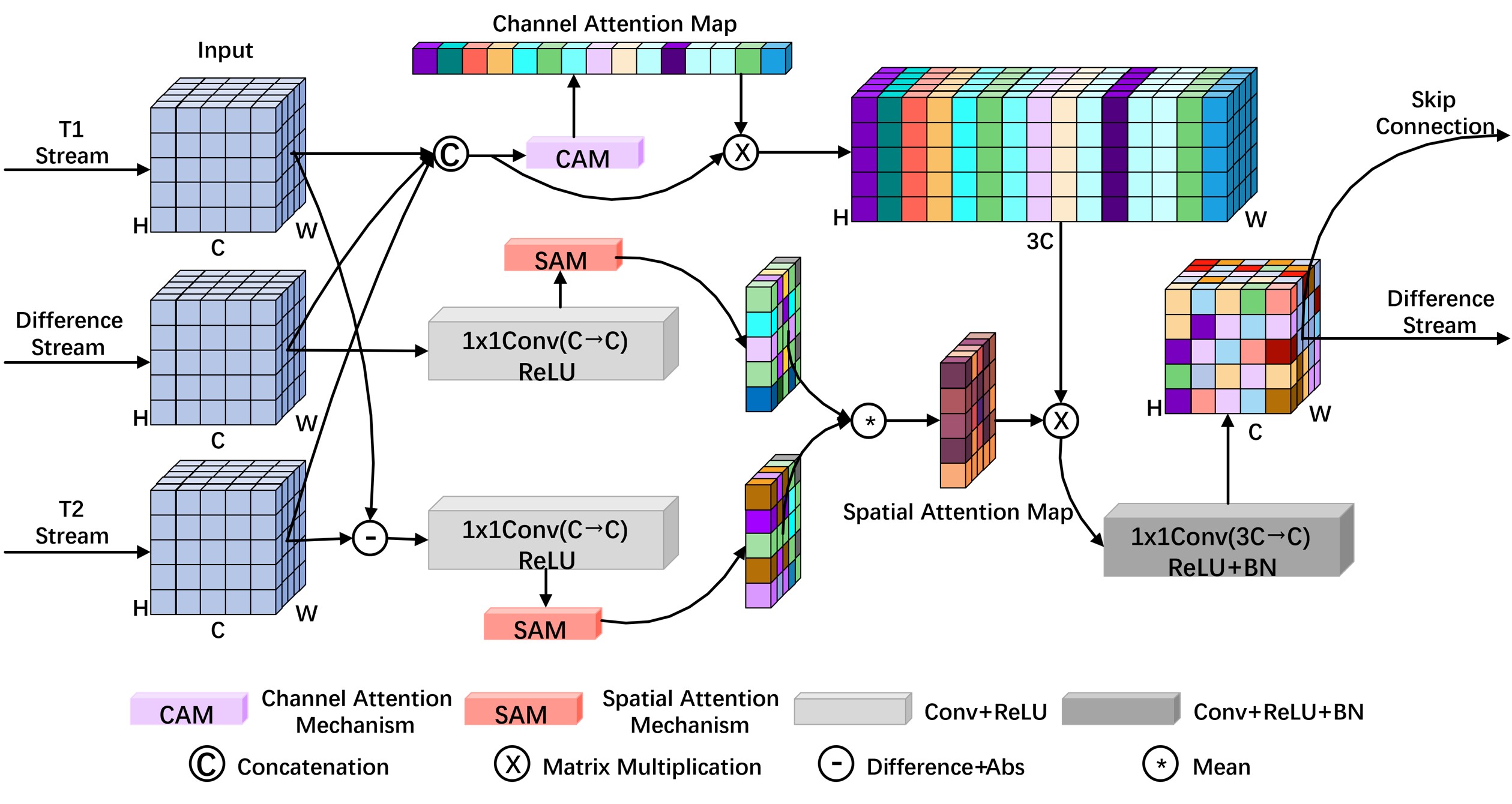}
	\caption{Structure of the MBSSCA module.}
	\label{fig3}
\end{figure*}

The structure of the MBSSCA module is illustrated in Fig. \ref{fig3}. The module first integrates and discriminates features from different branches in the channel and spatial dimensions separately. This is followed by an interaction fusion of features that are change-enhanced in different dimensions to obtain a new feature map with the help of the spatial-spectral cross attention. The fusion process of the multi-branch features in the channel dimension is shown in Eq. \ref{eq6}.
\begin{equation}
	\label{eq6}
	F_{MBSSCA_p}^{CAM}={\left[l_{p,Q}^{I_1},l_{p,Q}^{I_D},l_{p,Q}^{I_2}\right]}\times CAM\left(\left[l_{p,Q}^{I_1},l_{p,Q}^{I_D},l_{p,Q}^{I_2}\right]\right)
\end{equation}

where $[$·$,$·$,$·$]$ represents the concatenation; CAM represents the processing of the channel attention module. The structure and procedure of CAM are elaborated in Part D. $F_{MBSSCA_p}^{CAM}$ represents the feature map obtained by channel-wise fusion in the p-th layer of the MBSSCA module. We first concatenate the features of the three branches input to the p-th layer MBSSCA module, $l_{p,Q}^{I_1}$, $l_{p,Q}^{I_D}$, $l_{p,Q}^{I_2}$, and then utilize CAM to learn the channel-wise attention distribution feature vector related to the change target. Where the magnitude of each component of the feature vector is positively correlated with the relevance of the corresponding channel to detecting changes. The more valuable the channel information is for detecting changes, the larger the corresponding value of the vector component, and vice versa. Multiplying the obtained channel-wise attention distribution feature vector by the concatenated features of the three branches yields the channel-wise fusion feature $F_{MBSSCA_p}^{CAM}$, which achieves channel information filtering.

As for spatial dimension, we first difference the features $l_{p,Q}^{I_1}$,$l_{p,Q}^{I_2}$ of the branch T1 and T2 to obtain the difference intensity map of the deep features. Then, we perform a 1x1 convolution operation on the intensity map to increase its nonlinearity and enhance the robustness of the features. We utilize SAM to recognize and enhance the real change objects in the convolved features, with a spatial attention distribution feature map obtained. The procedure is formalized as follows.
\begin{equation}
	\label{eq7}
	W_{MBSSCA_p}^{SAM_{1,2}}=SAM\left(ReLU\left(Conv_{1\times1}\left(\left| l_{p,Q}^{I_1}-l_{p,Q}^{I_2}\right|\right)\right)\right)
\end{equation}

where $|$·$|$ represents the calculation of the absolute value, $Conv_{1×1}($·$)$ represents the 1x1 convolution operation, $ReLU($·$)$ represents the activation function. SAM represents the processing of the spatial attention module, the structure and processing of SAM are elaborated in Part D. $W_{MBSSCA_p}^{SAM_{1,2}}$ represents the spatial attention distribution feature map obtained from the spatial dimensional identification of the changed features in both T1 and T2 branches in the p-th layer MBSSCA module. Similarly, the corresponding spatial attention distribution feature map $W_{MBSSCA_p}^{SAM_{D}}$ is obtained by performing the same operations on the input features $l_{p,Q}^{I_D}$ in the TD branch, and the procedure is illustrated in Eq. \ref{eq8}.
\begin{equation}
	\label{eq8}
	W_{MBSSCA_p}^{SAM_D}=SAM\left(ReLU\left(Conv_{1\times1}\left(l_{p,Q}^{I_D}\right)\right)\right)
\end{equation}

For $W_{MBSSCA_p}^{SAM_{1,2}}$ and $W_{MBSSCA_p}^{SAM_{D}}$, the weights of the changed region pixels in the spatial attention distribution feature map are increased, while those of the unchanged region pixels are decreased. The original input of the TD branch is derived from directly subtracting the bi-temporal images. Though this process efficiently extracts the change information of the bi-temporal images, it may also introduce noise and pseudo-changes. These noise and pseudo-changes can be recognized and suppressed through deep semantic feature extraction, but there are still some stubborn pseudo-changes that are difficult to distinguish. And they appear as larger weights in the spatial attention distribution feature map. To address this issue, we compare the deep features of different levels extracted from the bi-temporal images to obtain discriminative features in a more abstract feature space. The features can focus on deeper semantic information, ignore scattered pseudo-change information caused by noise interference, and appear as smaller weights in the spatial attention distribution feature map. Through averaging the weights of the two spatial attention distribution feature maps, the weights of the pixels in the real change regions are relatively increased, while those in the unchanged regions are relatively decreased. Crucially, the weights of the pseudo-change region pixels are significantly reduced, effectively identifying and suppressing the pseudo-changes in the TD branch, thus achieving the goal of correcting the TD branch. The procedure is shown in Eq. \ref{eq9}.
\begin{equation}
	\label{eq9}
	W_{MBSSCA_p}^{SAM}=\frac{W_{MBSSCA_p}^{SAM_{1,2}}+W_{MBSSCA_p}^{SAM_D}}{2}
\end{equation}

Taking into account the weights of the two spatial attention distribution feature maps, we obtain a new feature map, denoted as $W_{MBSSCA_p}^{SAM}$. By multiplying it with the feature $F_{MBSSCA_p}^CAM$ obtained by fusing the features of the three branches in the channel dimension, we can interactively identify and enhance the change information of the three branches from the spectral dimension and the spatial dimension, thereby obtaining the fused features. As the number of channels of the fused features is tripled, there exists more redundant information. This not only increases the learning difficulty of the model, but also significantly affects the training efficiency. To address this, the 1x1 convolution is used to process the fused features, preserving significant information while filtering out the redundant information to achieve dimensionality reduction. The procedure of feature fusion is shown in Eq. \ref{eq10}.
\begin{equation}
	\begin{aligned}
		\label{eq10}
		&l_{MBSSCA_p}^{I_D}=\\
		&ReLU\left(BN\left(Conv_{1\times1}\left(W_{MBSSCA_p}^{SAM}\!\times\!{F_{MBSSCA_p}^{CAM}}\right)\right)\right)
	\end{aligned}
\end{equation}

Eq. \ref{eq6}-\ref{eq10} is the expansion of Eq. \ref{eq4}. The features from the three branches are interfused by the p-th layer MBSSCA module to finally yield a new fused feature $l_{MBSSCA_p}^{I_D}$. The fused features are then used as input to the TD branch for further deep feature extraction. In addition, the fused features are passed to the decoder stage through a skip-connection structure for change information supplementation at different coarse and fine granularities and for change detection.

\subsection{Decoder}

As introduced in Part B, the feature $l_{MBSSCA_5}^{I_D}$ is finally obtained by triplet encoder, which is used for decoder to further mine and recover to the resolution of the original image. Additionally, the change feature maps $l_{MBSSCA_p}^{I_D}(p=1,2,3,4,5)$, obtained from the MBSSCA module at different levels of the TD branch encoder, are transmitted to the corresponding levels of the decoder through skip-connections to achieve complementary fusion of different coarse- and fine-grained change features. In turn, the shallow and deep differential information is fully extracted and mined to effectively identify change objects of different sizes. Through CAM and SAM, the information such as shallow detail texture and deep semantic localization is fully extracted and integrated, and the fine-grained change information is supervised to be filtered and complementarily fused with the coarse-grained change information. By continuously enhancing the representation of true change features and weakening the representation of pseudo-change and unchanged features, comprehensive and accurate change features are provided for the detection of change, further improving the accuracy and performance of CD.

\begin{figure}[!t]
	\centering
	\includegraphics[width=3.4in]{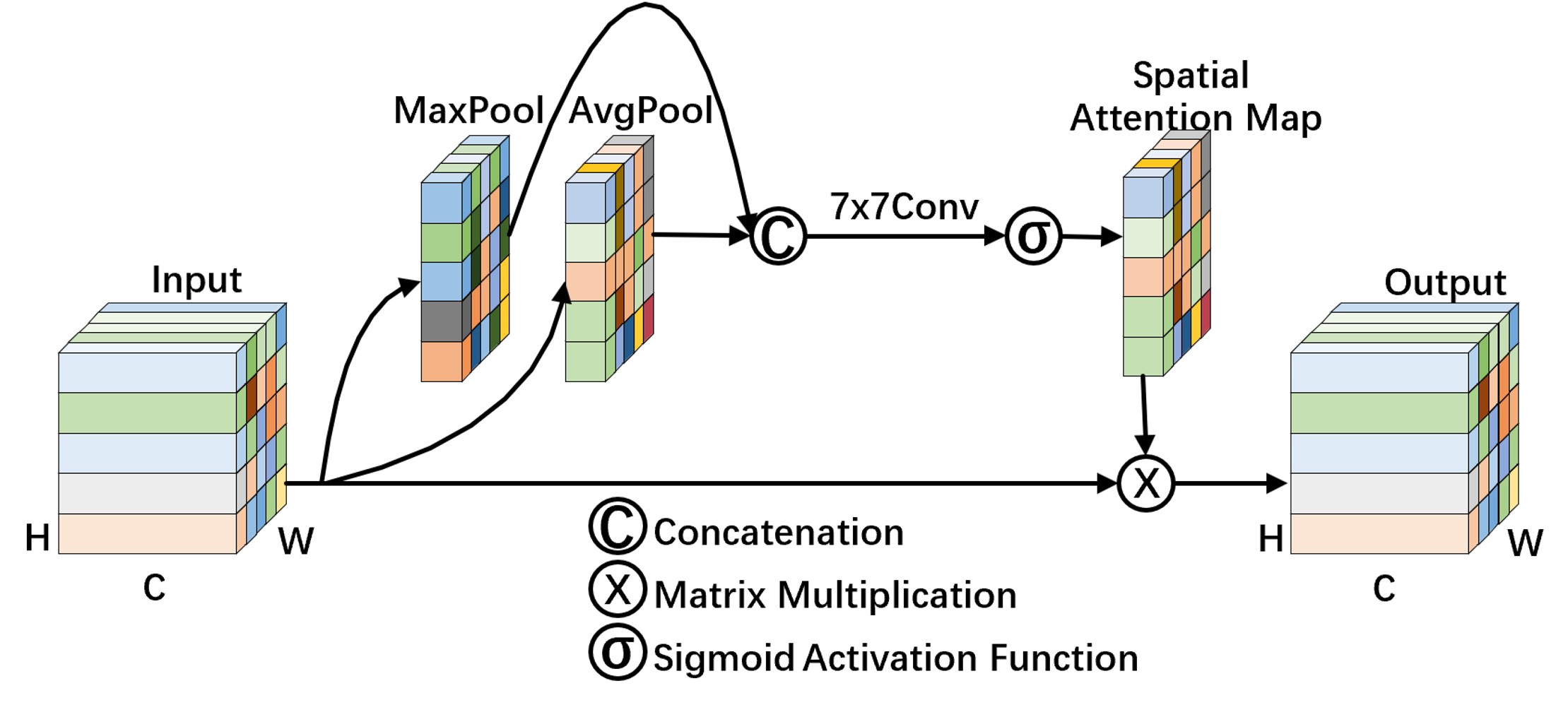}
	\caption{Structure of spatial attention module (SAM)}
	\label{fig4}
\end{figure}

As shown in Fig. \ref{fig1}, the decoder of the proposed T-UNet adopts a structure symmetric to the convolutional module in the triplet encoder as its backbone, which is mainly composed of 5 convolutional modules. Each convolutional module consists of 2 to 3 convolutional blocks, and each convolutional block consists of a convolutional layer, a batch normalization layer, and a ReLU activation function layer. In addition, the decoder also includes 5 spatial attention modules, 4 transpose convolutional layers, 4 feature concatenation layers, and 4 channel attention modules. After each convolutional module, the spatial attention module is used to identify the real change region in the feature map. The spatial attention focuses on the spatial feature information of each pixel, and the importance of the position of each pixel is encoded in the spatial attention distribution feature map. The weight of the position of each pixel in the spatial attention distribution feature map is automatically recalibrated during network training by repeatedly receiving feedback from the ground truth. By multiplying the input feature with the corresponding spatial attention distribution weights, the position of the pixels in the change region is assigned higher importance, while the position of the pixels in the unchanged region is assigned lower importance, thus enabling the network to learn to focus on the pixels in the change region that are critical for change detection from complex fused data. The structure of SAM is shown in Fig. \ref{fig4}. For the input feature, the global information of the feature is first obtained through max pooling and average pooling, and then the spatial attention distribution feature map is learned using convolutional operation. Finally, the spatial attention distribution feature map is multiplied with the input feature to obtain a new feature with enhanced attention in the change region.

The procedure of SAM is illustrated in Eq. \ref{eq11}.
\begin{equation}
	\begin{aligned}
		\label{eq11}
		\begin{cases}
			&W_{decoder_m}^{SAM}=\sigma\left(Conv_{7\times7}\left(\right.\right.\\
			&\left.\left.\left[AvgPool\left(l_{decoder_m}^{I_D}\right)\!,\! MaxPool\left(l_{decoder_m}^{I_D}\right)\right]\right)\right)\\
			&l_{{SAM}_m}^{I_D}=W_{decoder_m}^{SAM}\!\times\! l_{decoder_m}^{I_D}
		\end{cases}
	\end{aligned}
\end{equation}
where $l_{decoder_m}^{I_D}$ represents the output feature of the m-th convolutional module in the decoder, which is also the input feature of the m-th SAM. $AvgPool($·$)$ and $MaxPool($·$)$ represent the average pooling and max pooling operation along the channel axis, respectively. $[$·$]$ represents the concatenation of features. $Conv_{7×7}($·$)$ represents the 7x7 convolution operation, $\sigma($·$)$ represents the sigmoid activation function, $W_{decoder_m}^{SAM}$ represents the attention distribution feature map learned by the m-th SAM, and $l_{SAM_m}^{I_D}$ represents the output feature of the m-th SAM in the decoder. 

The change-enhanced features are obtained through the SAM module, and then the resolution of the features is gradually expanded by transpose convolutional operations. After the final transpose convolutional layer, the feature map is restored to the same size as the original bi-temporal images. The feature transformation of the transpose convolutional layer is illustrated in Eq. \ref{eq12}.
\begin{equation}
	\label{eq12}
	l_{CT_m}^{I_D}=ConvTranspose_m\left(l_{SAM_m}^{I_D}\right)
\end{equation}
where $ConvTranspose_m($·$)$ represents the m-th layer transpose convolutional operation in the decoder, with a kernel size of 2×2. $l_{CT_m}^{I_D}$ represents the output feature of the m-th layer transpose convolution. 

Through the operation of transpose convolution, the spatial size of the feature is doubled. Subsequently, it is concatenated with the feature passed from the MBSSCA module at the corresponding level of the triplet encoder to obtain the fused features $l_{concat_m}^{I_D}$, as shown in Eq. \ref{eq13}.
\begin{equation}
	\begin{aligned}
		\label{eq13}
		\begin{cases}
			l_{concat_m}^{I_D}&=\left[l_{CT_m}^{I_D},l_{MBSSCA_p}^{I_D}\right]\\
			m+p&=5
		\end{cases}
	\end{aligned}
\end{equation}
where $l_{MBSSCA_p}^{I_D}$ represents the output feature of the p-th layer MBSSCA module in the triplet encoder. As only the output features of the first four MBSSCA modules are skip-connected, so $p$ ranges from 1 to 4. $l_{CT_m}^{I_D}$ represents the output feature of the m-th layer transpose convolution, and similarly, m here takes values from 1 to 4. To deeply fuse fine-grained and coarse-grained change information, CAM is used to filter channel information after each concatenation layer. The CAM module focuses on spectral information related to the target of the CD task. The importance of each channel is encoded in the channel attention distribution feature map, and its weights are automatically recalibrated during network training. By multiplying the input features of the CAM module with the corresponding attention distribution weights, channels relevant to change information are enhanced and those not relevant to change information are suppressed. In this way, the network can focus on spectral information that is important for the target task from complex fusion data. 

\begin{figure}[!t]
	\centering
	\includegraphics[width=3.4in]{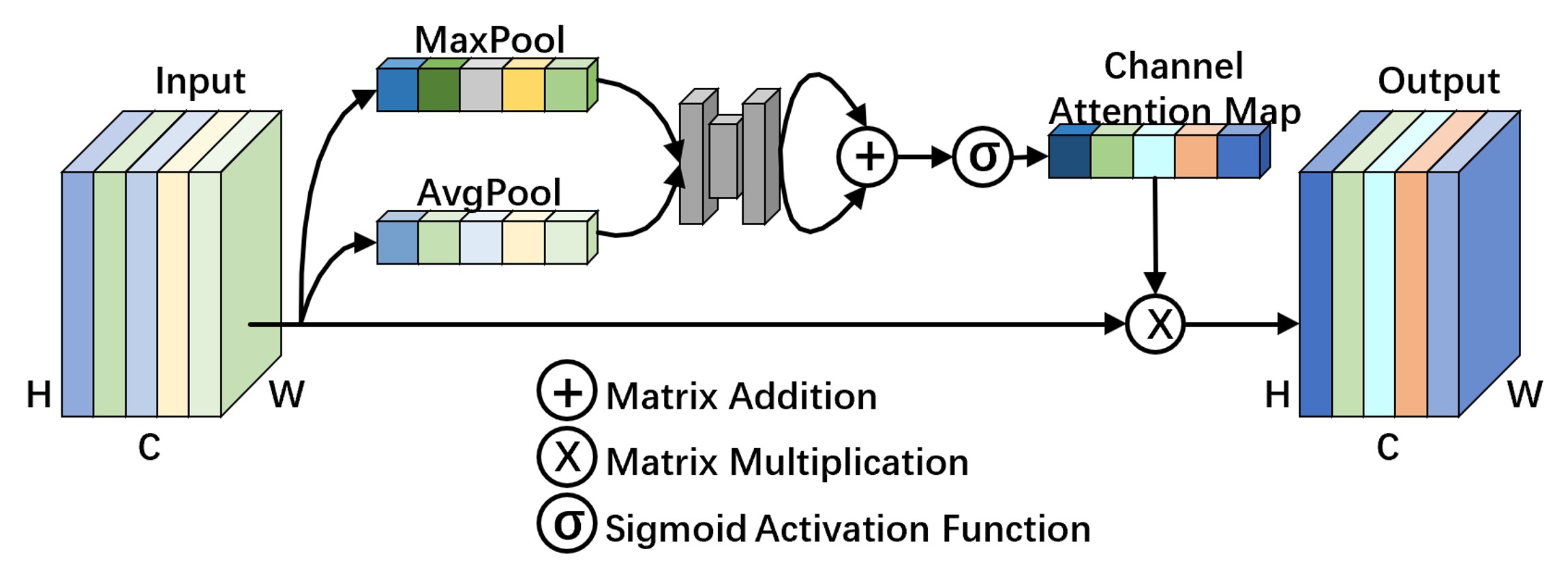}
	\caption{Structure of channel attention module (CAM)}
	\label{fig5}
\end{figure}

The structure of CAM is shown in Fig. \ref{fig5}. For the input feature $l_{concat_m}^{I_D}$, the spatial information is first compressed by the average pooling and the max pooling along the spatial axis. The two resulting vectors are then input into a shared multi-layer perceptron (MLP) to calculate the attention distribution weights. The outputs of MLP are merged using element-wise addition. The merged vector is then rectified using the sigmoid function, yielding an attention distribution feature vector $W_{decoder_m}^{CAM}$ assigned to each spectral channel. Finally, the channel attention distribution feature vector is multiplied with the input feature to obtain a new feature $l_{CAM_m}^{I_D}$.

The procedure of CAM is illustrated in Eq. \ref{eq14}.
\begin{equation}
	\begin{aligned}
		\label{eq14}
		\begin{cases}
			W_{decoder_m}^{CAM}&=\sigma\left(MLP\left(AvgPool\left(l_{concat_m}^{I_D}\right)\right)\right.\\
			&\left.+MLP\left(MaxPool\left(l_{concat_m}^{I_D}\right)\right)\right)\\
			l_{{CAM}_m}^{I_D}&=W_{decoder_m}^{CAM}\!\times\! l_{concat_m}^{I_D}
		\end{cases}
	\end{aligned}
\end{equation}

By filtering and focusing on the channel information through the m-th layer CAM, the feature $l_{CAM_m}^{I_D}$ is obtained, which is served as the input of the (m+1)-th layer convolutional module to further mining the change features and gradually recovering to the original image size. At the end of the decoder, the sigmoid function is employed to activate the features output from the last layer to obtain the change probability map. The final change detection result is obtained by thresholding.

\subsection{Loss Function}

In the experiments, we use a joint loss function based on the sigmoid binary cross-entropy loss and the dice loss, as illustrated in Eq. \ref{eq15}.
\begin{equation}
	\label{eq15}
	L_{total}=L_{sigmoid\_bce}+L_{dice}
\end{equation}

The binary cross-entropy loss is illustrated in Eq. \ref{eq16}, commonly used in problems of binary classification. The goal of our CD task is to classify the changed category from the unchanged category, which can be treated as a binary classification problem as well.
\begin{equation}
	\label{eq16}
	L_{bce}=\sum\nolimits_i\left(-y_ilog\left(\hat{y_i}\right)-\left(1-y_i\right)log\left(1-\hat{y_i}\right)\right)
\end{equation}
where $i\in[1,N]$, N represents the total number of pixels in the sample, $y_i$ represents the true label, and $\hat{y_i}$ represents the predicted label. However, as the value of $\hat{y_i}$ approaches zero, the value of $log(\hat{y_i})$ will converge to infinity, resulting in an infinitely large loss that severely damages the training of the model. Similarly, as the value of $\hat{y_i}$ converges to 1, the value of $log(\hat{y_i})$ will tend to be infinitesimal, resulting in an infinite loss, which also interferes with the training of the model. 

To solve this problem, we use the sigmoid function to first process the predicted values to make the predicted values converging to 0 and 1 towards the middle, alleviating the problem of infinity loss. The expression of the sigmoid binary cross-entropy loss is illustrated in Eq. \ref{eq17}.
\begin{equation}
	\label{eq17}
	L_{sigmoid\_bce}\!=\!\sum\nolimits_i\left(-y_ilog\left(\sigma\left(\hat{y_i}\right)\right)\!-\!\left(1\!-\! y_i\right)log\left(\sigma\left(1\!-\!\hat{y_i}\right)\right)\right)
\end{equation}
where $\sigma($·$)$ represents the sigmoid function. Moreover, the change detection task suffers from a severe sample imbalance. The number of changed pixels is much smaller than that of unchanged pixels.

In addition, the change detection task suffers from a severe sample imbalance, with the number of changed pixels being much smaller than the number of unchanged pixels. The training of the model will be biased towards the category with a large number of samples, making the loss insensitive to the category with fewer samples. To address the sample imbalance issue, we train the model in combination with the dice loss. The dice coefficient is commonly used to evaluate the similarity of two sets, and here we employ it to focus on the similarity of the set of change regions. The expression of the dice loss is illustrated in Eq. \ref{eq18}.
\begin{equation}
	\label{eq18}
	L_{dice}\!=\!1\!-\!\frac{2\left|Y\cap\hat{Y}\right|}{|Y|+|\hat{Y}|}\!=\!1\!-\!\frac{2\sum\nolimits_iy_y\hat{y_i}}{\sum\nolimits_iy_i+\sum\nolimits_i\hat{y_i}}
\end{equation}
where $Y$ represents the set of true labels and $\hat{Y}$ represents the set of predicted labels. When the true label is the unchanged category, $y_i$ equals 0, the value of $y_i\hat{y_i}$ is 0, so that the calculation of the corresponding dice loss will not be activated. When the true label is the changed category, $y_i$ equals 1 and the calculation of dice loss will be activated to achieve the purpose of focusing on the change regions. 

\section{Experimental Setting}
\label{Experimental Setting}
To evaluate the performance of the proposed network T-UNet, experiments were carried out on three publicly available datasets, LEVIR-CD, WHU-CD and DSIFN-CD. Additionally, seven state-of-the-art CD models were selected for comparison. All models were trained and tested on an NVIDIA GeForce RTX 3090 GPU with 24 GB of memory and were implemented using the DL framework Pytorch. In the experiments, the Adam optimization algorithm with an initial learning rate of 1e-4 is used to update the weights of the model parameters. The following describes the datasets used in the experiments, the compared methods and the evaluation metrics in turn.

\subsection{Data Description}
Three publicly available CD datasets are employed in this paper for the evaluation of the model performance, namely the LEVIR-CD dataset, the WHU-CD dataset and the DSIFN-CD dataset.

\begin{figure*}[!t]
	\centering
	\includegraphics[width=6in]{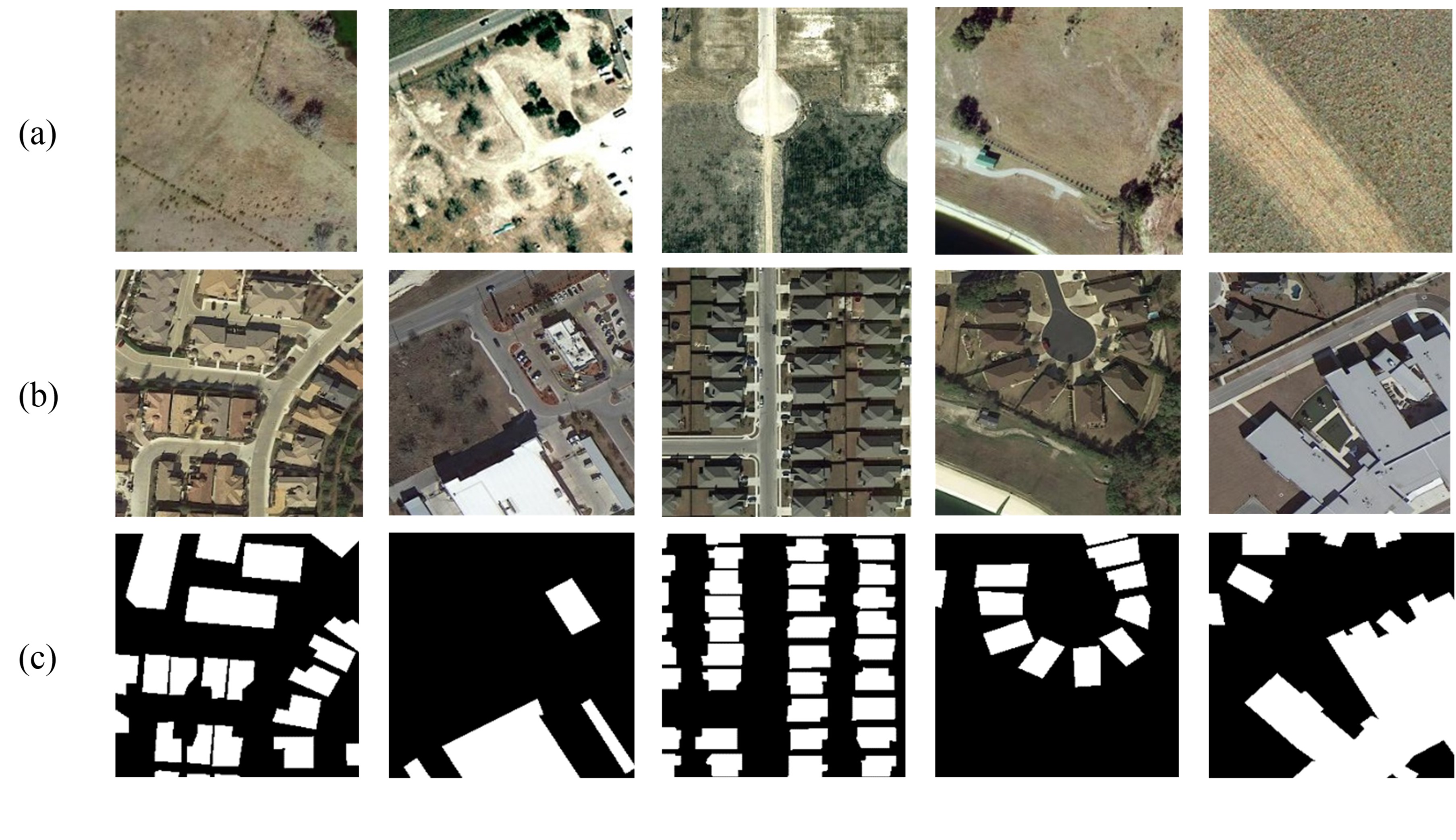}
	\caption{Some samples from the LEVIR-CD dataset. (a) T1 image (b) T2 images (c) Label.}
	\label{fig6}
\end{figure*}

1) LEVIR-CD dataset \citep{ref52}: this dataset is concerned with changes related to buildings, including the occurrence of buildings (changes from soil/grass/hardened ground/building under construction to completed buildings) and the destruction of buildings. The LEVIR-CD dataset consists of 637 sets of bi-temporal high-resolution RS image pairs of size 1024*1024, with a spatial resolution of 0.5m/pixel. Since the size of the original images is too large, using them directly for network training will greatly enlarge the demand on the graphic memory and increase the training time. Therefore, in this paper, the original images are cropped in 256*256 block to obtain 10192 sets of high-resolution RS image pairs with a size of 256*256. The dataset contains a total of 31,333 individual building change instances, with an average of 3 building changes per image pair. In this paper, the dataset is randomly divided in the ratio of 7:1:2, where the training set contains 7120 image pairs, the validation set contains 1024 image pairs, and the remaining 2048 image pairs are used as the test set. Some samples in the LEVIR-CD dataset are shown in Fig. \ref{fig6}.

\begin{figure*}[!t]
	\centering
	\includegraphics[width=6in]{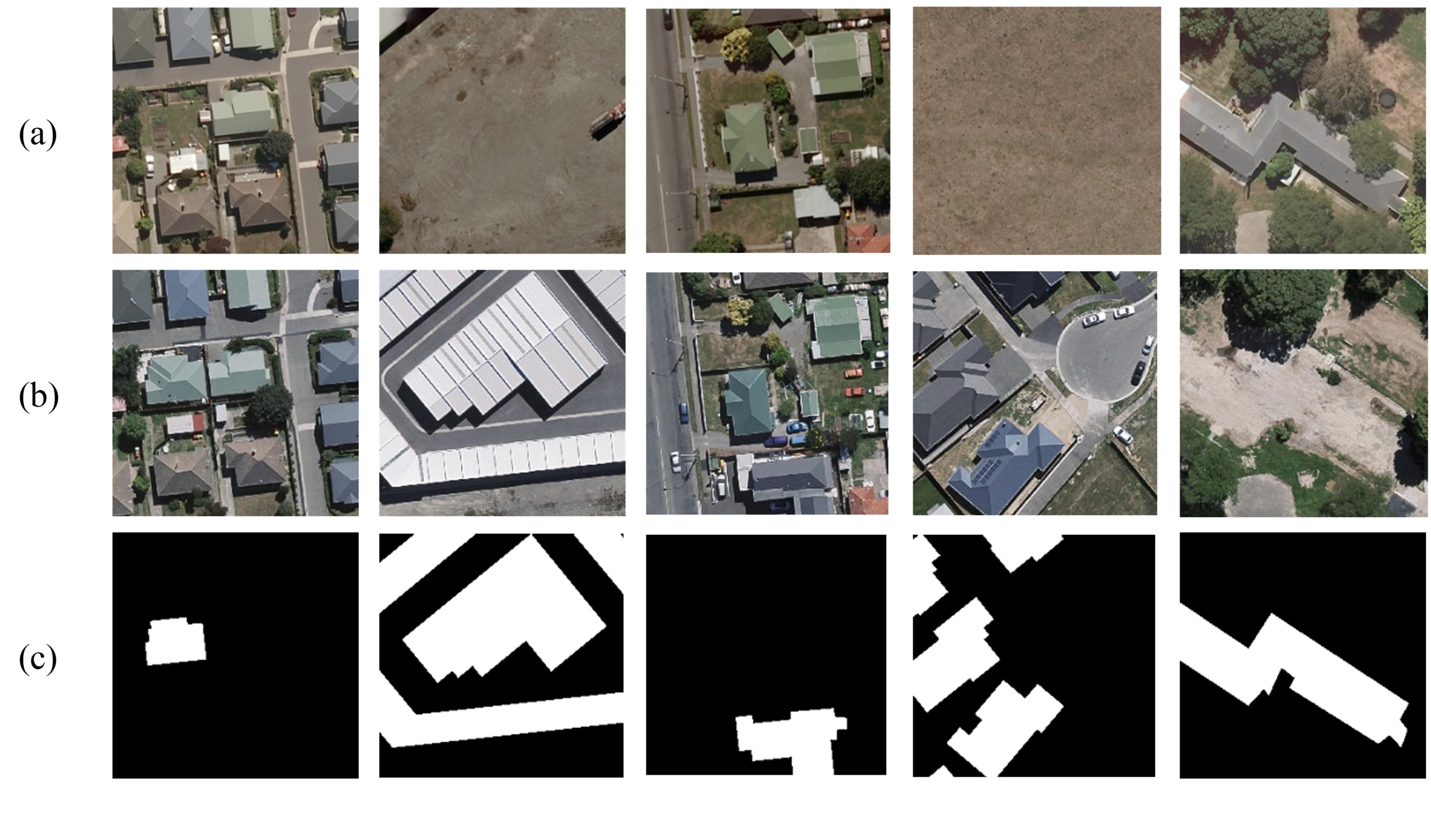}
	\caption{Some samples from the WHU-CD dataset. (a) T1 image (b) T2 images (c) Label.}
	\label{fig7}
\end{figure*}

2) WHU-CD dataset \citep{ref53}: this dataset is also focused on building changes. It contains one set of bi-temporal high-resolution RS image pairs of size 32507*15354, with a spatial resolution of 0.3m/pixel. Similarly, we perform a block cropping operation on the original images with a size of 256*256, and finally obtain 7620 sets of bi-temporal high-resolution RS image pairs with a size of 256*256. In this paper, the dataset is randomly divided in the ratio of 8:1:1, where the training set contains 6096 image pairs, the validation set contains 762 image pairs, and the remaining 762 image pairs are used as the test set. Some samples in the WHU-CD dataset are shown in Fig. \ref{fig7}.

\begin{figure*}[!t]
	\centering
	\includegraphics[width=6in]{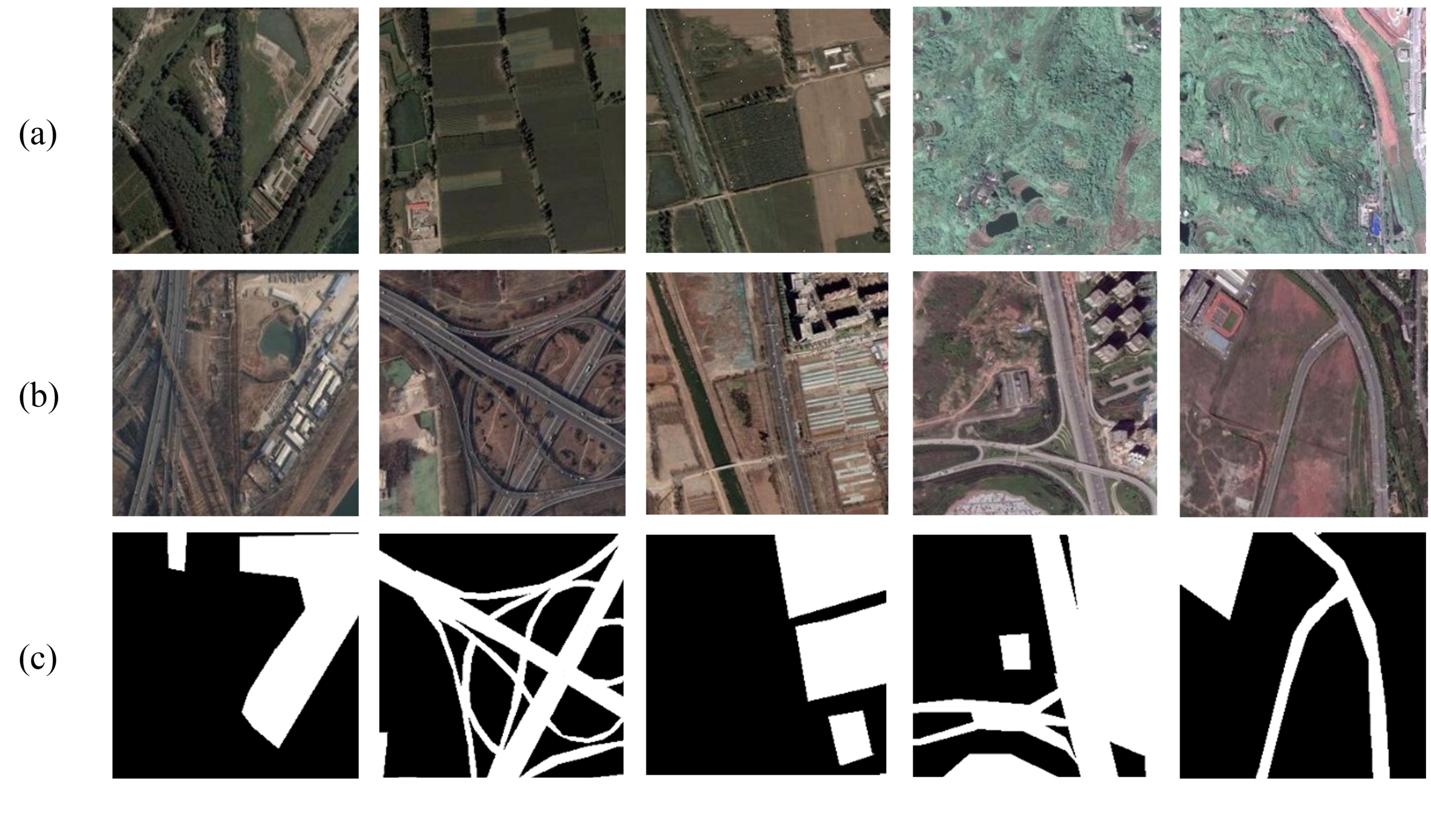}
	\caption{Some samples from the DSIFN-CD dataset. (a) T1 image (b) T2 images (c) Label.}
	\label{fig8}
\end{figure*}

3) DSIFN-CD dataset \citep{ref45}: this dataset consists of six sets of co-registrated bi-temporal high-resolution RS image pairs with a spatial resolution of 2m/pixel. The six image pairs were acquired from six cities in China, including Beijing, Shenzhen, Chengdu, Chongqing, Wuhan and Xi'an. The authors of the dataset first cropped the image data from the first five cities (Beijing, Shenzhen, Chengdu, Chongqing and Wuhan) to obtain 394 sets of image pairs of 512*512 in size. The data was then expanded to 3940 sets by a series of data augmentation operations such as rotation, flipping and random cropping, of which 3600 sets were randomly selected as the training set and the remaining 340 sets were used as the validation set. The image data of Xi'an was directly cropped into 192 sets of image pairs with a size of 256*256, and all these data were used as the test set. Finally, 15952 sets of bi-temporal high-resolution RS image pairs of size 256*256 were obtained, of which the training set contained 14400 image pairs, the validation set contained 1360 image pairs, and the remaining 192 image pairs were used as the test set. Some samples in the DSIFN-CD dataset are shown in Fig. \ref{fig8}.

The three datasets differ in terms of data composition and division, so they have different emphases when evaluating the models. The LEVIR-CD dataset and the WHU-CD dataset are designed to train and detect models with greater robustness by introducing pseudo-changes such as seasonal variation and light variation. The training and test sets were divided in the DSIFN-CD dataset using image data from different cities, whereas the external conditions (climate, geographical location, radiation conditions, etc.) of the images acquired in different locations can affect the final imaging results. This makes training and testing on the DSIFN-CD dataset extremely challenging for the generalization and transfer capability of the model.

\subsection{Compared Methods}
To validate the superiority of the proposed method, seven state-of-the-art CD methods were selected for comparative testing in this paper. A brief description of these methods is as follows.

1) FC-EF \citep{ref43}: Network FC-EF is constructed on the basis of UNet, with only one branch in its encoder. The EF structure in the network FC-EF first concatenates the bi-temporal images, and then inputs them into an end-to-end network for training. It draws inspiration from the idea of skip connection in UNet, where feature maps of the same size from the down-sampling process are concatenated with those from the up-sampling process, thus leveraging shallow local edge and texture information to supplement lost detail information during global semantic feature extraction.

2) FC-Siam-Conc \citep{ref43}: Network FC-Siam-Conc is an extension of network FC-EF, which uses a Siamese structure containing two encoders with the identical structure and shared weights. Unlike network FC-EF, the feature fusion of the bi-temporal images occurs during feature extraction in the Siamese structure, rather than before inputting into the network.

3) FC-Siam-Diff \citep{ref43}: Network FC-Siam-Diff is another variant of network FC-EF. Similar to network FC-Siam-Conc, it uses a Siamese structure containing two encoders with the identical structure and shared weights. However, network FC-Siam-Diff takes the difference between the feature maps of the two branches in the down-sampling process, and then passes them to the corresponding up-sampling layer for concatenation, the purpose of which is to highlight the difference information in the bi-temporal images.

4) DASNet \citep{ref51}: Network DASNet uses a Siamese structure with two branches in its encoder. It employs dual attention modules to extract new features that contain long-range contextual information in both spatial and channel domains. Weighted double-margin contrastive loss is used to reduce the distance between pixels in unchanged regions and increase the distance between pixels in changed regions, thus enhancing the real change features.

5) SNUNet \citep{ref49}: Network SNUNet uses a Siamese subnetwork as the encoder to extract deep features from the original bi-temporal images, with two branches in its encoder. An ensemble channel attention module is used to aggregate and refine features of multiple semantic levels extracted from the nested U-Net with dense connections, which suppresses semantic gaps and localization errors to some extent.

6) DSIFN \citep{ref45}: Network DSIFN uses a Siamese structure for feature extraction from the bi-temporal images. Attention mechanism is introduced to enhance features relevant to the target task when fusing them. The deep supervision module is embedded in the middle layer of the network to directly back-propagate the error of prediction, effectively addressing the problem of gradient disappearance in deep network structures.

7) DESSN \citep{ref50}: Network DESSN is also constructed based on the Siamese UNet structure. The difference enhancement module is employed at each feature extraction layer to process the difference map of the features extracted from the bi-temporal images. A weighted difference map is obtained and then added to the input features to enhance the concentration on real change feature information and edge details in the changed region. Meanwhile, the enhanced difference map is used for skip connection, providing rich and accurate change feature information for the decoder.

\subsection{Evaluation Metrics}

In the experiments, four widely used evaluation metrics, including Overall Accuracy (OA), Precision (Pre), Recall (Rec), and F1-Score (F1), are adopted to quantitatively evaluate the performance of each model for change detection in high-resolution remote sensing images. The changed pixels are defined as positive samples and the unchanged pixels as negative samples. The relationship among the parameters TP, TN, FN, and FP is shown in Table \ref{tab1}.

\begin{table}[!t]
	\centering
	\caption{Relation of parameters TP, TN, FN and FP}
	\label{tab1}
	
	\begin{tabular}{|p{4.19em}|p{4.19em}|p{4.19em}|}
		\toprule
		\multirow{2}[4]{*}{Real value} & \multicolumn{2}{p{8.38em}|}{Predictive value} \\
		\cmidrule{2-3}    \multicolumn{1}{|c|}{} & Positive & Negative \\
		\midrule
		Positive & TP    & FN \\
		Negative & FP    & TN \\
		\bottomrule
	\end{tabular}
	
\end{table}

1) OA: OA is the proportion of the number of correctly predicted pixels in the test to the total number of pixels tested, and the formula for calculating OA is shown in Eq. \ref{eq19}.
\begin{equation}
	\label{eq19}
	OA=\frac{TP+TN}{TP+TN+FN+FP}
\end{equation}

2) Pre: Pre refers to the proportion of correctly predicted positive samples to the total number of samples predicted as positive. In the experiments, we define the changed pixels as positive samples and the unchanged pixels as negative samples. For the task of change detection, a higher precision indicates fewer false alarms, i.e., fewer unchanged pixels being predicted as changed. The higher the value, the more correct predictions are made. The formula for calculating Pre is shown in Eq. \ref{eq20}.
\begin{equation}
	\label{eq20}
	Pre=\frac{TP}{TP+FP}
\end{equation}

3) Rec: Rec refers to the proportion of correctly predicted positive samples in the total number of positive samples in the test dataset. For change detection tasks, a higher recall rate indicates fewer missed detections, i.e., fewer false negatives, which means that fewer changed pixels are predicted as unchanged pixels. A higher recall indicates that more predictions are correct. The calculation formula for Rec is shown in Eq. \ref{eq21}.
\begin{equation}
	\label{eq21}
	Rec=\frac{TP}{TP+FN}
\end{equation}

4) F1: Pre and Rec are a set of contradictory metrics, where high precision is usually associated with low recall and high recall is associated with low precision. Therefore, F1 is adopted to evaluate Pre and Rec, providing a more comprehensive and accurate evaluation of experimental results. F1 represents the harmonic mean of Pre and Rec, and its calculation formula is shown in Eq. \ref{eq22}.
\begin{equation}
	\label{eq22}
	F1=\frac{2\times Pre\times Rec}{Pre+Rec}
\end{equation}

\section{Experimental Results and Discussion}
\label{Experimental Results and Discussion}
In this section, we first detail and analyze the experimental results of the proposed T-UNet with other comparative models on three datasets, then investigate the relationship between model complexity and accuracy, and finally verify the effectiveness of the proposed modules through ablation studies.

\subsection{Experiments on the LEVIR-CD Dataset}
\label{sec_4_1}

The evaluation metrics of different models on the LEVIR-CD dataset are shown in Table  \ref{tab2}. The model FC-Siam-Conc achieves the highest recall (96.74\%) among all models, which is comparable to the other two models of the same scale, FC-EF (95.79\%) and FC-Siam-Diff (91.83\%), but far exceeds several other models with more complex structures. However, the precision of models FC-EF (35.01\%), FC-Siam-Conc (48.56\%), and FC-Siam-Diff (57.09\%) is the lowest among all the models, indicating that these three models cannot effectively distinguish between real change factors and pseudo-change factors such as seasonal and lighting changes, resulting in few missed detections but many false alarms. 

On the basis of these three CD models, DASNet, SNUNet, DSIFN, and DESSN all introduce attention mechanisms in different manners, effectively concentrating attention on the change area and ignoring redundant information that affects change discrimination, thus further improving the accuracy of CD task. DASNet (F1: 85.09\%, OA: 98.43\%) introduces a dual attention module in the Siamese structure of the encoding stage, increasing the distance between pixels in the change area and reducing the distance between pixels in the unchanged area to effectively identify the change area. SNUNet (F1: 89.22\%, OA: 98.92\%) combines dense connection modules and ECAM to aggregate and refine semantic features at different levels, suppressing problems such as loss of deep localization information and positioning errors in the neural network, so as to extract enough accurate change feature information for the final decision. The deep feature fusion module of DSIFN (F1: 89.48\%, OA: 98.95\%) introduces SAM and CAM to effectively enhance the features in the change region and suppress the redundant information during the fusion phase. The DE module of DESSN (F1: 90.06\%, OA: 99.01\%) employs an attention mechanism to obtain a weighted difference map at each feature extraction layer, which is added to the original features to further enhance attention on the change area. Meanwhile, using the enhanced difference map for skip connections provides richer and more accurate change feature information for the decoding stage. However, these four attention networks are all based on the Siamese structure and more or less ignore some of the differences between original bi-temporal images in deep feature extraction. The triple-branch network T-UNet proposed in this paper integrates both original image features and difference information, and continuously corrects the difference information through the MBSSCA module to detect changes. The edge details and texture information of change objects between bi-temporal images are fully identified, further reducing false alarms and missed detections. Compared with all comparative methods, T-UNet achieves the best results in Pre (92.60\%), F1 (91.63\%), and OA (99.16\%). The proposed T-UNet improves F1 by 1.57\%-40.35\%, demonstrating the excellent CD performance of T-UNet.

\begin{table}[!t]
	\centering
	\caption{Evaluation indexes of each change detection model on the LEVIR-CD dataset}
	\label{tab2}
	
	\begin{tabular}{ccccc}
		\toprule
		Model & Pre(\%) & Rec(\%) & F1(\%) & OA(\%) \\ 
		\midrule
		FC-EF & 35.01 & 95.79 & 51.28 & 90.72 \\ 
		FC-Siam-Conc & 48.56 & \textbf{96.40} & 64.59 & 94.61 \\ 
		FC-Siam-Diff & 57.09 & 91.83 & 70.41 & 96.06 \\ 
		DASNet & 82.81 & 87.49 & 85.09 & 98.43 \\ 
		SNUNet & 91.23 & 87.29 & 89.22 & 98.92 \\ 
		DSIFN & 89.43 & 89.52 & 89.48 & 98.95 \\ 
		DESSN & 92.59 & 87.66 & 90.06 & 99.01 \\ 
		T-Unet & \textbf{92.60} & 90.68 & \textbf{91.63} & \textbf{99.16} \\
		\bottomrule
	\end{tabular}
	
\end{table}

\begin{figure*}[!t]
	\centering
	\includegraphics[width=7in]{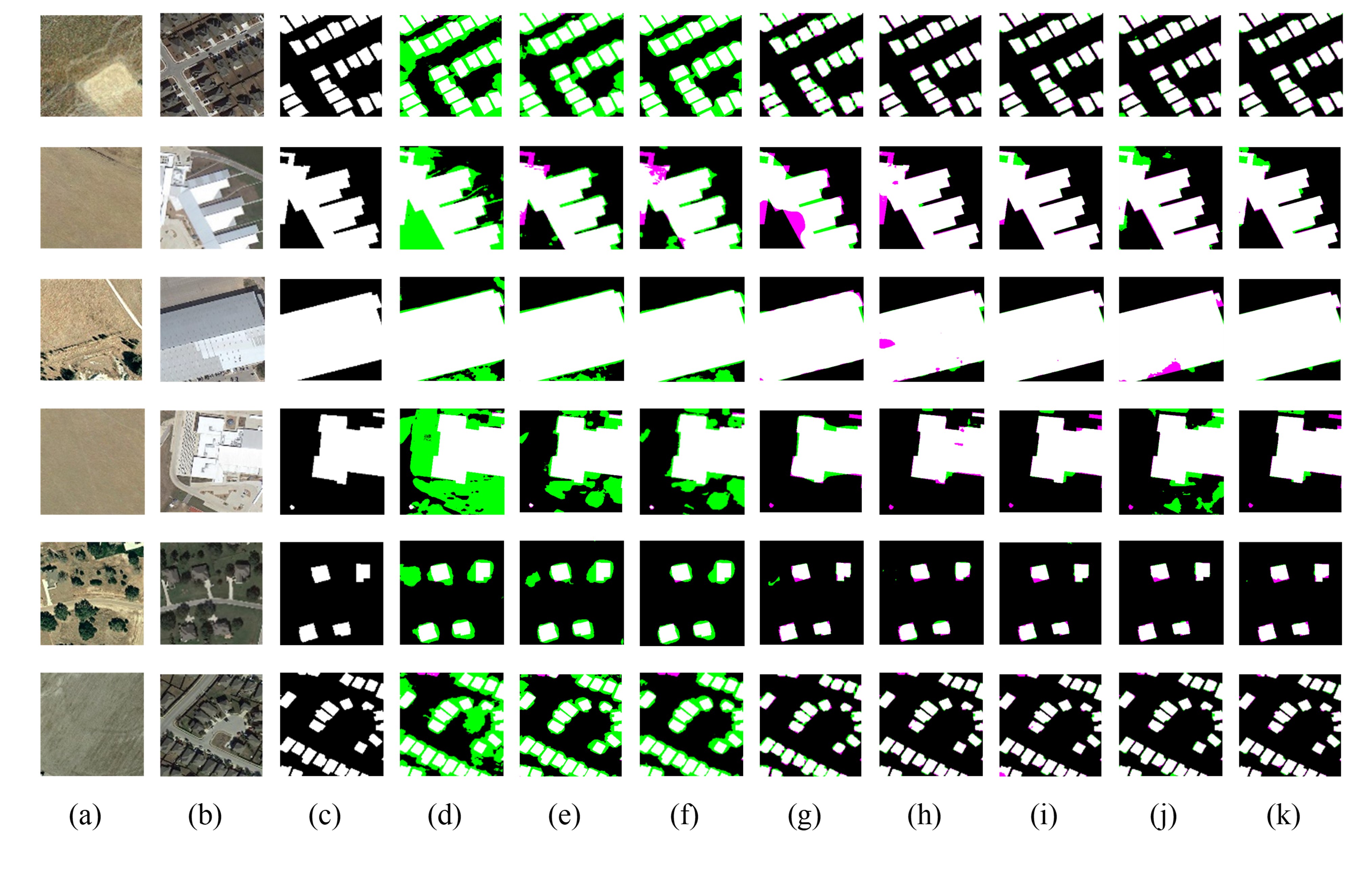}
	\caption{Visualization of change detection results on the LEVIR-CD dataset (a) T1 image; (b) T2 image; (c) Label; (d) FC-EF; (e) FC-Siam-Conc; (f) FC-Siam-Diff; (g) DASNet; (h) SNUNet; (i) DSIFN; (j) DESSN; (k)T-UNet. Green denotes false alarm, and purple denotes missed detection.}
	\label{fig9}
\end{figure*}

To further evaluate the change detection performance of different models on the LEVIR-CD dataset, the test results of all the compared models are visualized and compared with the original bi-temporal images as well as the labels, as shown in Fig. \ref{fig9}. As can be seen from the figure, all models can well capture the building change instances, but the performance of different models in recognizing the boundaries of change areas varies. The visualization results of models FC-EF, FC-Siam-Conc, and FC-Siam-Diff (figure \ref{fig9}, columns d-f) show that there are many false alarms around each building change instance, especially between small and dense change instances, which matches the high recall and low precision in Table \ref{tab2}. Models DASNet, SNUNet, DSIFN, and DESSN (figure \ref{fig9}, columns g-j) enhance the focus on change regions by introducing attention mechanisms, which greatly reduces the proportion of false alarms, but there are still some false alarms and missed detections in the discrimination of the edge details of the changed objects. It is noteworthy that the visualization results of the proposed T-UNet (figure \ref{fig9}, column k) have the most accurate boundary of the change area, which are closest to the label. This indicates that the proposed T-UNet can well mine and fuse the difference information of the original bi-temporal images and their difference image through the three-branch structure, accurately locate the changed objects and identify their edge details, thus effectively improving the accuracy of the proposed model.

\subsection{Experiments on the WHU-CD Dataset}
\label{sec_4_2}

The evaluation metrics results of different models tested on the WHU-CD dataset are shown in Table \ref{tab3}. Similar to the results obtained on the LEVIR-CD dataset, the proposed T-UNet achieved the best results in Pre (95.44\%), F1 (91.77\%) and OA (99.42\%). The model with the highest recall is FC-Siam-Conc (95.45\%), while the recall rates of the other two models with the same scale, FC-EF (93.60\%) and FC-Siam-diff (91.75\%), are only slightly lower than that of FC-Siam-Conc, significantly higher than the other comparable models. However, the precision rates of models FC-EF (35.01\%), FC-Siam-Conc (48.56\%), and FC-Siam-diff (57.09\%) are the lowest among all models, indicating that these three models have few misses but many false alarms. Based on these three CD models, DASNet (F1: 87.97\%, OA: 99.11\%) introduces a dual attention module to enhance the recognition of change regions. SNUNet (F1: 89.38\%, OA: 99.23\%) effectively aggregates and refines semantic features at different levels using dense connection modules and ECAM mechanism, mining rich and accurate change information for change detection. DSIFN (F1: 88.57\%, OA: 99.15\%) introduces two attention mechanisms in the feature fusion stage to enhance features of the change area and suppress redundant channel information respectively. DESSN (F1: 88.78\%, OA: 99.19\%) uses the DE module to compare the features of the two branches in the encoder for mining change details, while transmitting shallow change information to the decoder for complementary fusion of features at different granularities. Compared with models FC-EF, FC-Siam-Conc, and FC-Siam-Diff, these four attention networks based on the Siamese structure have greatly improved in the evaluation metrics of Pre, F1, and OA, indicating that the attention mechanism plays a crucial role. The proposed three-branch network T-UNet comprehensively utilizes features of the original bi-temporal images and their difference image, fully excavating edge detail texture information of real changed objects between bi-temporal images, and further reducing false alarms and misses. Considering all evaluation metrics, the proposed T-UNet performs the best. Compared with other comparative models, the evaluation metric of F1 has been improved by 2.39\%-39.82\%.

\begin{table}[!t]
	\centering
	\caption{Evaluation indexes of each change detection model on the WHU-CD dataset}
	\label{tab3}
	
	\begin{tabular}{ccccc}
		\toprule
		Model & Pre(\%) & Rec(\%) & F1(\%) & OA(\%) \\ 
		\midrule
		FC-EF & 35.95 & 93.60 & 51.95 & 93.74 \\ 
		FC-Siam-Conc & 45.54 & \textbf{95.45} & 61.66 & 95.70 \\ 
		FC-Siam-Diff & 55.85 & 91.75 & 69.43 & 97.07 \\ 
		DASNet & 86.74 & 89.23 & 87.97 & 99.11 \\ 
		SNUNet & 89.48 & 89.28 & 89.38 & 99.23 \\ 
		DSIFN & 86.51 & 90.73 & 88.57 & 99.15 \\ 
		DESSN & 89.60 & 87.98 & 88.78 & 99.19 \\ 
		T-Unet & \textbf{95.44} & 88.37 & \textbf{91.77} & \textbf{99.42} \\ 
		\bottomrule
	\end{tabular}
	
\end{table}

\begin{figure*}[!t]
	\centering
	\includegraphics[width=7in]{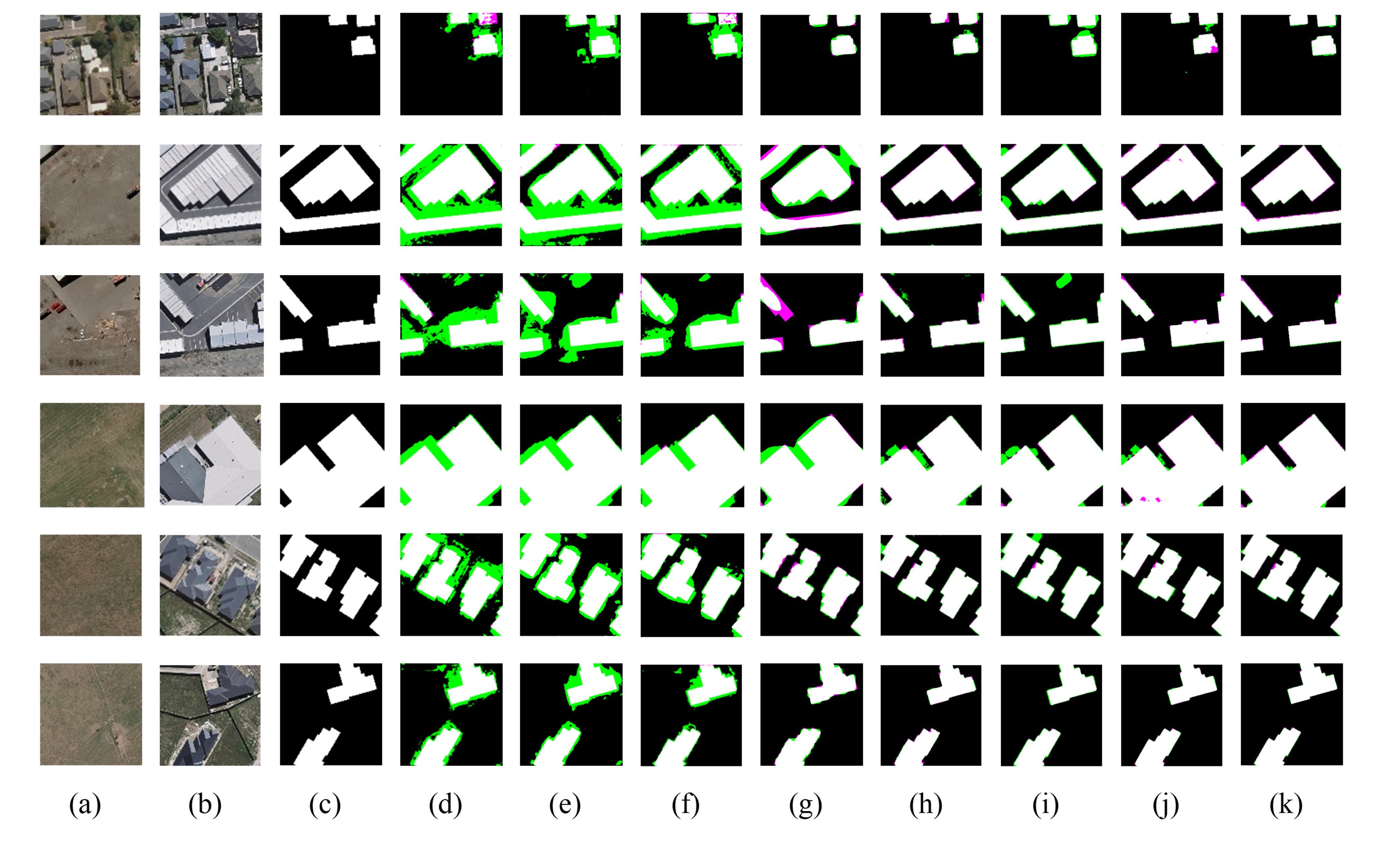}
	\caption{Visualization of change detection results on the WHU-CD dataset (a) T1 image; (b) T2 image; (c) Label; (d) FC-EF; (e) FC-Siam-Conc; (f) FC-Siam-Diff; (g) DASNet; (h) SNUNet; (i) DSIFN; (j) DESSN; (k)T-UNet. Green denotes false alarm, and purple denotes missed detection.}
	\label{fig10}
\end{figure*}

To further evaluate the change detection performance of different models on the WHU-CD dataset, the test results of all the compared models are visualized and compared with the original bi-temporal images as well as the labels, as shown in Fig. \ref{fig10}. Similar to the LEVIR-CD dataset, there are many connections and blurred boundaries between the change instances in the visualization results of models FC-EF, FC-Siam-Conc, and FC-Siam-diff (figure \ref{fig10}, columns d-f), i.e. many false alarms, consistent with the quantitative results in Table \ref{tab3}. DASNet (figure \ref{fig10}, column g) introduces the dual attention mechanism to enhance the identification of change regions and reduce the false alarm rate. However, DASNet directly uses the deep features extracted from bi-temporal images to calculate the Euclidean distance as the criterion for discriminating changes, which cannot effectively utilize the change information, resulting in the loss of details of change instances and significant errors in the discrimination of the edges. Models SNUNet, DSIFN, and DESSN (figure \ref{fig10}, columns h-j) utilize the attention mechanism in the feature fusion stage of the bi-temporal images to effectively mine the difference features and fuse them, further reducing the false alarm rate and improving the edge details of the change instances. Compared with other state-of-the-art models, the proposed T-UNet (figure \ref{fig10}, column k) can fully exploit the features of the original bi-temporal images and their differential change features, comprehensively consider the spatial-spectral information to identify and detect the weak change features at the edge of change instances accurately. As can be seen from Fig. \ref{fig10}, among all the visualization results, the proposed T-UNet has the most accurate boundaries of the changed areas, which is closest to the label.

\subsection{Experiments on the DSIFN-CD Dataset}
\label{sec_4_3}

The evaluation metrics results of different models tested on the DSIFN-CD dataset are presented in Table \ref{tab4}. It can be noticed that FC-Siam-Conc has the highest recall (93.96\%), but its precision (34.76\%), F1 (50.75\%), and OA (69.01\%) are the lowest among all models, indicating that the model fails to distinguish real change from pseudo-change, resulting in a large number of false alarms. The model FC-EF using a single-branch encoder structure is also unable to extract enough key deep change features for change detection, with evaluation metric results comparable to those of the model FC-Siam-Conc. FC-Siam-Diff repeatedly takes the difference of features between the bi-temporal images for change feature extraction, while the model DASNet utilizes a Siamese neural network to extract deep features and then calculates their Euclidean distance as the discriminative feature for change detection, both models achieving significant improvements in F1 (61.90\%, 59.52\%) and OA (83.97\%, 83.34\%). The models SNUNet, DSIFN, and DESSN all introduce attention mechanisms in different manners, effectively focusing attention on the change areas and ignoring redundant information that has an impact on change discrimination, further improving the accuracy of change detection. However, none of the models reached an OA of 90\% and an F1 of 70\%, indicating that all models have not effectively learned the feature representation and model weights representative of the overall dataset during training on the DSIFN-CD dataset, thus resulting in poor performance on the test dataset.

The training and testing data of DSIFN-CD dataset were obtained from different sources. The training set consists of bi-temporal images from five cities, namely Beijing, Chengdu, Shenzhen, Chongqing and Wuhan; while the testing set is composed of bi-temporal images from Xi'an. Various external conditions such as weather, geographic location and radiation when imaging at different locations may affect the final imaging results, hence posing a great challenge to the generalization ability of CD models trained and tested on data obtained from different regions. Overall, the proposed T-UNet shows the best change detection performance on the DSIFN-CD dataset, achieving the best results in terms of Pre (70.86\%), F1 (69.52\%) and OA (89.83\%) among all models. Compared to other benchmark models, the evaluation metric of F1 is improved by 4.41\%-18.77\%, with the lower bound of improvement being higher than the other two datasets. This on one hand indicates that T-UNet has the most significant improvement on DSIFN-CD dataset, and on the other hand, demonstrates that the proposed T-UNet has the best generalization ability as well as robustness among all models.

\begin{table}[!t]
	\centering
	\caption{Evaluation indexes of each change detection model on the DSIFN-CD dataset}
	\label{tab4}
	
	\begin{tabular}{ccccc}
		\toprule
		Model & Pre(\%) & Rec(\%) & F1(\%) & OA(\%) \\ 
		\midrule
		FC-EF & 38.38 & 82.58 & 52.40 & 74.51 \\ 
		FC-Siam-Conc & 34.76 & \textbf{93.96} & 50.75 & 69.01 \\ 
		FC-Siam-Diff & 51.93 & 76.61 & 61.90 & 83.97 \\ 
		DASNet & 50.70 & 72.07 & 59.52 & 83.34 \\ 
		SNUNet & 60.77 & 68.24 & 64.29 & 87.11 \\ 
		DSIFN & 58.14 & 73.99 & 65.11 & 86.52 \\ 
		DESSN & 54.31 & 75.63 & 63.22 & 85.04 \\ 
		T-Unet & \textbf{70.86} & 68.23 & \textbf{69.52} & \textbf{89.83} \\
		\bottomrule
	\end{tabular}
	
\end{table}

\begin{figure*}[!t]
	\centering
	\includegraphics[width=7in]{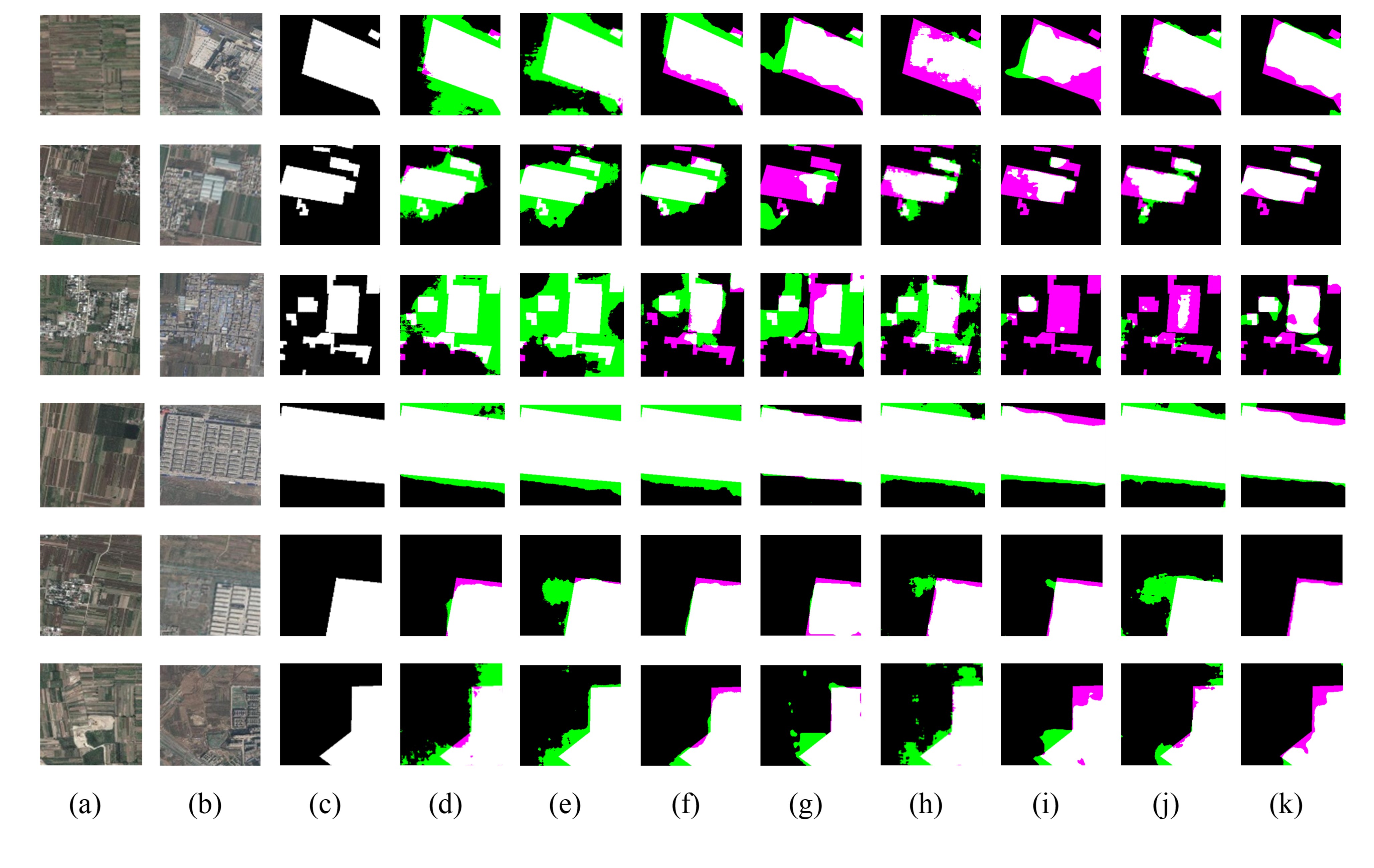}
	\caption{Visualization of change detection results on the DSIFN-CD dataset (a) T1 image; (b) T2 image; (c) Label; (d) FC-EF; (e) FC-Siam-Conc; (f) FC-Siam-Diff; (g) DASNet; (h) SNUNet; (i) DSIFN; (j) DESSN; (k)T-UNet. Green denotes false alarm, and purple denotes missed detection.}
	\label{fig11}
\end{figure*}

To further evaluate the change detection performance of different models on the DSIFN-CD dataset, the test results of all the compared models are visualized and compared with the original bi-temporal images as well as the labels, as shown in Fig. \ref{fig11}. For small and compact change instances (figure \ref{fig11}, rows 2-3), models FC-EF, FC-Siam-Conc, and FC-Siam-Diff cannot distinguish the boundaries between different change instances and detect them as a larger whole. The models DASNet, DSIFN, and DESSN missed multiple small change instance targets. The model SNUNet can locate individual change instances, but the boundaries between them are still not accurately determined. The proposed T-UNet is able to accurately locate each change instance and distinguish their boundaries, performing the best among all models. For large change instances (figure \ref{fig11}, rows 1, 4-6), all models can locate the positions of change instances, but with varying degrees of deficiency in the discrimination of boundaries. The models FC-EF, FC-Siam-Conc, and FC-Siam-Diff detect a large number of unchanged pixels as changed pixels, making the change instances in the result maps larger than the true change instances. The models DASNet, SNUNet, DSIFN, and DESSN reduce the false alarm rate but greatly increase the missed detection rate, making some of the change instances in the result maps smaller than the true change instances. The proposed T-UNet effectively balances the false alarm and missed detection cases among comparative models by fully utilizing the original image features and their differential change features. Among all the visualisation results, the proposed T-UNet shows the most accurate change area boundaries and is closest to the true labels, demonstrating excellent change detection capability.

\subsection{Complexity Analysis}

\begin{table*}[!t]
	\centering
	\caption{Model complexity metrics versus comprehensive evaluation metrics on each dataset}
	\label{tab5}
	
\begin{tabular}{cccccccccc}
	\toprule
	\multirow{2}[2]{*}{Model} & \multirow{2}[2]{*}{Params.(M)} & \multirow{2}[2]{*}{FLOPs(G)} & \multirow{2}[2]{*}{FLOPs/Params(K)} & \multicolumn{2}{c}{LEVIR-CD} & \multicolumn{2}{c}{WHU-CD} & \multicolumn{2}{c}{DSIFN-CD} \\
	&       &       &       & OA(\%)    & F1(\%)    & OA(\%)    & F1(\%)    & OA(\%)    & F1(\%) \\
	\midrule
	FC-EF & \textbf{1.35 } & \textbf{3.13 } & 2.32  & 90.72 & 51.28 & 93.74 & 51.95 & 74.51 & 52.40 \\
	FC-Siam-Conc & 1.55  & 4.89  & 3.15  & 94.61 & 64.59 & 95.70  & 61.66 & 69.01 & 50.75 \\
	FC-Siam-Diff & \textbf{1.35 } & 4.29  & 3.18  & 96.06 & 70.41 & 97.07 & 69.43 & 83.97 & 61.90 \\
	DASNet & 16.26  & 56.92  & 3.50  & 98.43 & 85.09 & 99.11 & 87.97 & 83.34 & 59.52 \\
	SNUNet & 3.01  & 11.79  & 3.92  & 98.92 & 89.22 & 99.23 & 89.38 & 87.11 & 64.29 \\
	DSIFN & 35.99  & 79.04  & 2.20  & 98.95 & 89.48 & 99.15 & 88.57 & 86.52 & 65.11 \\
	DESSN & 18.35  & 27.10  & \textbf{1.48 } & 99.01 & 90.06 & 99.19 & 88.78 & 85.04 & 63.22 \\
	T-Unet & 53.47  & 96.90  & 1.81  & \textbf{99.16} & \textbf{91.63} & \textbf{99.42} & \textbf{91.77} & \textbf{89.83} & \textbf{69.52} \\
	\bottomrule
\end{tabular}%
	
\end{table*}

To compare the performance of the models in a comprehensive and fair manner, we evaluate the complexity of the proposed T-UNet and other models in terms of the number of parameters (Params.) and the floating-point operations (FLOPs). \ref{tab5} displays the complexity metrics of the different models and their comprehensive evaluation metrics on each dataset. Among them, FLOPs/Params. is introduced to evaluate the effect of the increase in the number of parameters on model computation. The smaller the value of FLOPs/Params., the smaller the effect of the increase in parameters on the increase in model computation and vice versa.

As shown in \ref{tab5}, both Params. and FLOPs of the model FC-EF are at the minimum level. In addition, the comprehensive evaluation metrics OA and F1 of FC-EF on all three datasets are also at the lowest level. This indicates that the structure of the model is so simple that it is difficult to learn deep features for effectively distinguishing changed ground object. The models FC-Siam-Conc and FC-Siam-Diff, which are similar in size to the model FC-EF, also show a low number of parameters, a low computational volume as well as a low model accuracy. Moreover, these three models have larger values of FLOPs/Params. indicating that the model computation corresponding to each unit of parameter is larger. With the introduction of the attention mechanism, the Params. and FLOPs of the proposed T-UNet are significantly larger with the other four comparison models. However, it is clear that the comprehensive evaluation metrics OA and F1 of these models on the three datasets are substantially improved. Among them, the models DSIFN and DESSN have lower FLOPs/Params. than the three previously mentioned models without the introduction of the attention mechanism. As the proposed T-UNet has one more branch for feature extraction compared to other models, T-UNet has the maximum value in both Params. and FLOPs. But it has the second smallest FLOPs/Params. of all the models, indicating that the model computation corresponding to each unit of parameter is small. That is, the increase in the number of parameters of T-UNet did not bring about an increase in the amount of model computation in the same proportion. It is worth noting that the proposed T-UNet achieves the maximum values of the comprehensive evaluation metrics OA and F1 on all three datasets while maintaining a reasonably acceptable number of Params. and FLOPs. On the whole, our analysis shows that the proposed T-UNet achieves a good tradeoff between complexity and performance, enabling it to serve as a practical choice for bi-temporal RS image CD tasks.

\subsection{Ablation Study}

To further validate the effectiveness of the proposed modules in the network T-UNet, a series of ablation experiments are conducted on the LEVIR-CD, WHU-CD and DSIFN-CD datasets. In the experiments, the same hyperparameters and details as those in the comparative experiments are kept. The ablation experiments consist of three parts, which are used to validate the effectiveness of the triplet encoder, MBSSCA module, and attention modules in the decoder, respectively. Table \ref{tab6} shows the results of the ablation experiments, where Triplet, MBSSCA, and AM represent the triplet encoder, MBSSCA module, and attention modules in the decoder of the corresponding network, respectively. The symbol "\checkmark" in the Table \ref{tab6} indicates that the module of the corresponding column is included in the model of the corresponding row. For example, the network Ours/6 contains the triplet encoder and MBSSCA module. Single and Siamese represent the number of branches in the encoder of the CD network, which are 1 and 2, respectively. Specifically, Single indicates that only the TD branch of difference image data flow is used for feature extraction and change detection, while Siamese indicates that a double-branch structure with the identical architecture and shared weights is employed for feature extraction and change detection, corresponding to the two branches T1 and T2 in T-UNet. The decoder backbone used in the ablation experiments is similar for all networks, and the only difference is the presence or absence of attention modules in the decoder, as shown in column AM of Table \ref{tab6}.

\begin{table*}
	\centering
	\caption{Results of ablation experiments}
	\label{tab6}
	
	\begin{tabular}{ccccccccc}
		\toprule
		Dataset & Model & Triplet & MBSSCA & AM    & Pre(\%) & Rec(\%) & F1(\%) & OA(\%) \\
		\midrule
		\multirow{8}[2]{*}{LEVIR-CD} & Single/1 &       &       &       & 70.99 & 78.21 & 74.42 & 97.26 \\
		& Single/2 &       &       & \checkmark     & 85.49 & 68.21 & 75.88 & 97.79 \\
		& Siamese/3 &       &       &       & 85.56 & 91.25 & 88.31 & 98.76 \\
		& Siamese/4 &       &       & \checkmark     & 91.39 & 89.31 & 90.33 & 99.02 \\
		& Ours/5 & \checkmark     &       &       & 87.71 & \textbf{92.97} & 90.26 & 98.97 \\
		& Ours/6 & \checkmark     & \checkmark     &       & 89.77 & 92.22 & 90.97 & 99.07 \\
		& Ours/7 & \checkmark     &       & \checkmark     & 90.12 & 91.63 & 90.86 & 99.06 \\
		& Ours/T-Unet & \checkmark     & \checkmark     & \checkmark     & \textbf{92.60}  & 90.68 & \textbf{91.63} & \textbf{99.16} \\
		\midrule
		\multirow{8}[2]{*}{WHU-CD} & Single/1 &       &       &       & 70.33 & 88.60  & 78.42 & 98.23 \\
		& Single/2 &       &       & \checkmark     & 88.90  & 74.33 & 80.97 & 98.73 \\
		& Siamese/3 &       &       &       & 81.73 & \textbf{91.85} & 86.49 & 98.96 \\
		& Siamese/4 &       &       & \checkmark     & 89.71 & 86.23 & 87.94 & 99.14 \\
		& Ours/5 & \checkmark     &       &       & 90.89 & 88.49 & 89.67 & 99.26 \\
		& Ours/6 & \checkmark     & \checkmark     &       & 92.03 & 89.72 & 90.86 & 99.34 \\
		& Ours/7 & \checkmark     &       & \checkmark     & 95.14 & 87.21 & 91.00    & 99.37 \\
		& Ours/T-Unet & \checkmark     & \checkmark     & \checkmark     & \textbf{95.44} & 88.37 & \textbf{91.77} & \textbf{99.42} \\
		\midrule
		\multirow{8}[2]{*}{DSIFN-CD} & Single/1 &       &       &       & 51.43 & 61.40  & 55.98 & 83.58 \\
		& Single/2 &       &       & \checkmark     & 53.07 & 61.80  & 57.11 & 84.22 \\
		& Siamese/3 &       &       &       & 58.83 & 65.79 & 62.11 & 86.36 \\
		& Siamese/4 &       &       & \checkmark     & 67.32 & 59.55 & 63.19 & 88.21 \\
		& Ours/5 & \checkmark     &       &       & 63.87 & 66.23 & 65.03 & 87.51 \\
		& Ours/6 & \checkmark     & \checkmark     &       & 60.80  & \textbf{74.63} & 67.01 & 87.89 \\
		& Ours/7 & \checkmark     &       & \checkmark     & \textbf{72.26} & 61.69 & 66.56 & 89.46 \\
		& Ours/T-Unet & \checkmark     & \checkmark     & \checkmark     & 70.86 & 68.23 & \textbf{69.52} & \textbf{89.83} \\
		\bottomrule
	\end{tabular}
	
\end{table*}

1) Discussion on the effectiveness of triplet encoder: Triplet encoder is proposed to extract and utilize the original object features and their differential change features between the pre- and post-time phases, reducing the loss of key difference information. To analyze the performance of the triplet encoder, we compare the single-branch network Single/1 that processes difference images, and the dual-branch network Siamese/3 that processes the bi-temporal images, with the proposed triple-branch network Ours/5 containing the triplet encoder that simultaneously processes bi-temporal images and their difference images. As shown in Table \ref{tab6}, the F1 results of network Ours/5 on the LEVIR-CD, WHU-CD, and DSIFN-CD datasets are respectively improved by 1.95\%-15.84\%, 3.18\%-11.25\%, and 2.92\%-9.05\% compared to networks Single/1 and Siamese/3. It is worth noting that all evaluation metrics of network Ours/5 on the LEVIR-CD and DSIFN-CD datasets are better than those of networks Single/1 and Siamese/3. This indicates that the proposed triplet encoder can simultaneously extract the differential change information of the single branch and the features of the pre- and post-time phases, effectively integrating their complementary information to improve the detection accuracy of the change area, and thus verifying the effectiveness of the proposed triplet encoder.

2) Discussion on the effectiveness of the attention modules in the decoder: In the decoder stage of the proposed T-UNet, we introduce CAM and SAM to facilitate feature fusion and feature filtering, and to enhance channel and spatial features related to changes. To analyze the performance of the attention modules in the decoder, we embed the same structure of CAM and SAM into the decoder of the networks Single/1, Siamese/3, and Ours/5, and obtain the corresponding attention networks Single/2, Siamese/4, and Ours/7. As shown in Table \ref{tab6}, with the addition of the attention modules, network Single/2 shows a significant improvement in F1 and OA compared to network Single/1 on the three datasets, by 1.13\%-2.55\% and 0.50\%-0.64\%, respectively. Similarly, network Siamese/4 improves F1 and OA by 1.08\%-2.02\% and 0.18\%-1.85\% respectively over the three datasets compared to network Siamese/3. Network Ours/7 improves F1 and OA by 0.60\%-1.53\% and 0.09\%-1.95\%, respectively, compared with network Ours/5 on the three datasets. It is indicated that the attention modules in the decoder can effectively assist the network to select useful information during feature fusion and enhance the representation of change-related features. This ultimately improves the performance of the CD networks and thus validates the effectiveness of the attention modules in the decoder of the proposed network T-UNet.

3) Discussion on the effectiveness of MBSSCA module: The MBSSCA module is proposed to interactively fuse information from the three branches of the triplet encoder, utilizing the contrast information of deep features from the bi-temporal images in the branch T1 and T2 to continuously correct difference information in the TD branch, thus detecting changes accurately. To evaluate the performance of the MBSSCA module, the pure triple-branch network Ours/5 and the triple-branch attention network Ours/7 are added the MBSSCA module to obtain network Ours/6 and the proposed network Ours/T-UNet for comparison. As shown in Table \ref{tab6}, with the addition of the MBSSCA module to the triplet encoder, network Ours/6 achieves significant improvements in F1 and OA on all three datasets compared to network Ours/5, with an increase of 0.71\%-1.98\% and 0.10\%-1.95\%, respectively. Similarly, the proposed network Ours/T-UNet also achieves significant improvements in F1 and OA on all three datasets compared to network Ours/7, with an increase of 0.77\%-2.96\% and 0.05\%-0.37\%, respectively. Notably, both network Ours/6 and Ours/T-UNet outperform network Ours/5 and Ours/7 on the WHU-CD dataset for all evaluation metrics, respectively. It is evident from the results that the network with the addition of the MBSSCA module shows a significant improvement in the overall evaluation metrics compared to the network without the MBSSCA module. This indicates that the MBSSCA module facilitates the interaction of information from different branches and the mining of change features, thus validating the effectiveness of the MBSSCA module.

We take the pure triple-branch network Ours/5 as the base and then gradually add each proposed module, i.e. MBSSCA module (Ours/6), attention module in decoder (Ours/7), and the combination of the MBSSCA module and attention module in decoder (Ours/T-UNet), for module validation. As shown in Table \ref{tab6}, the addition of each module improves the overall evaluation metrics for change detection. Notably, the proposed network Ours/T-UNet achieves the best performance in terms of F1 and OA on all three datasets. Compared to the baseline network Ours/5, as well as networks Ours/6 and Ours/7, the proposed network Ours/T-UNet achieves improvements of 1.37\%-4.49\% and 0.16\%-2.32\%, 0.66\%-2.51\% and 0.08\%-1.94\%, and 0.77\%-2.96\% and 0.05\%-2.37\% in terms of F1 and OA, respectively. These results demonstrate that the performance of the baseline is significantly improved with the cumulative addition of each proposed module. Again, this validates the effectiveness of the modules proposed in the paper.

\section{Conclusion}
\label{Conclusion}
In this paper, we propose a novel triple-branch encoder-based network framework for change detection. The network framework is capable of simultaneously extracting the original object features and their differential change features between the pre- and post-time phases. To effectively interact and fuse the feature information of the three branches in the triplet encoder, we propose a multi-branch spatial-spectral cross attention module (MBSSCA). In the decoder stage of the network, we introduce the channel attention mechanism (CAM) and spatial attention mechanism (SAM) to fully explore and integrate shallow details and textures, as well as deep semantic information. In this way, comprehensive and accurate change features are available to support the detection of changes. The proposed network T-UNet is compared with seven other state-of-the-art change detection networks on three publicly available datasets. Experimental results show that the proposed network T-UNet achieves the best results for the evaluation metrics F1 and OA on the LEVIR-CD, WHU-CD and DSIFN-CD datasets, significantly outperforming the other seven compared methods. It demonstrates that the proposed T-UNet can well identify and localize the true change features in the bi-temporal images, showing the superior change detection capability. However, the existing models still have some shortcomings, and in the future, we will further improve the change detection performance from the aspects of model generalization ability and the way of focusing on change information, thus meeting the needs of more complex scenarios.

\section*{Declaration of Competing Interest}
The authors declare that they have no known competing financial interests or personal relationships that could have appeared to influence the work reported in this paper. 

\section*{Acknowledgment}
This work was supported in part by the National Key Research and Development Program of China under Grant 2022YFB3903300 and 2022YFB3903405, and in part by the National Natural Science Foundation of China under Grant T2122014, 61971317, 62225113 and 42230108.

\bibliographystyle{cas-model2-names}
%
\bibliography{refs.bib}

%
%

\end{document}